\documentclass[10pt,twocolumn,letterpaper]{article}

\usepackage{iccv}
\usepackage{times}
\usepackage{epsfig}
\usepackage{graphicx}
\usepackage{amsmath}
\usepackage{amssymb}

\usepackage{multirow}
\usepackage{multicol}
\usepackage{tabularx} 

\newcommand{\SimpleClick}{\texttt{SimpleClick}}
\usepackage{enumitem}
\usepackage{cite}
\usepackage{pifont}
\newcommand{\cmark}{\ding{51}}
\newcommand{\xmark}{\ding{55}}
\usepackage{tabularx,colortbl}
\usepackage{tablefootnote}
\usepackage{stix}
\usepackage{booktabs}
\usepackage{appendix}

\usepackage[pagebackref=true,breaklinks=true,letterpaper=true,colorlinks,bookmarks=false]{hyperref}

\iccvfinalcopy 

\def\iccvPaperID{2457} 

\ificcvfinal\fi

\usepackage[capitalize]{cleveref}
\crefname{section}{Sec.}{Secs.}
\Crefname{section}{Section}{Sections}
\Crefname{table}{Table}{Tables}
\crefname{table}{Tab.}{Tabs.}

\begin{document}

\title{SimpleClick: Interactive Image Segmentation with Simple Vision Transformers}


\author{Qin Liu, Zhenlin Xu, Gedas Bertasius, Marc Niethammer \\
University of North Carolina at Chapel Hill \\
{\tt\small \href{https://github.com/uncbiag/SimpleClick}{https://github.com/uncbiag/SimpleClick}}}

\maketitle
\ificcvfinal\thispagestyle{empty}\fi

\begin{abstract}
Click-based interactive image segmentation aims at extracting objects with a limited user clicking. A hierarchical backbone is the \emph{de-facto} architecture for current methods. Recently, the plain, non-hierarchical Vision Transformer (ViT) has emerged as a competitive backbone for dense prediction tasks. This design allows the original ViT to be a foundation model that can be finetuned for downstream tasks without redesigning a hierarchical backbone for pretraining. Although this design is simple and has been proven effective, it has not yet been explored for interactive image segmentation. To fill this gap, we propose {\SimpleClick}, the first interactive segmentation method that leverages a plain backbone. Based on the plain backbone, we introduce a symmetric patch embedding layer that encodes clicks into the backbone with minor modifications to the backbone itself. With the plain backbone pretrained as a masked autoencoder (MAE), {\SimpleClick} achieves state-of-the-art performance. Remarkably, our method achieves \textbf{4.15} NoC@90 on SBD, improving \textbf{21.8\%} over the previous best result. Extensive evaluation on medical images demonstrates the generalizability of our method. We further develop an extremely tiny ViT backbone for {\SimpleClick} and provide a detailed computational analysis, highlighting its suitability as a practical annotation tool.
\end{abstract}

\section{Introduction}
\label{sec:intro}
The goal of interactive image segmentation is to obtain high-quality pixel-level annotations with limited user interaction such as clicking. Interactive image segmentation approaches have been widely applied to annotate large-scale image datasets, which drive the success of deep models in various applications, including video understanding~\cite{xu2018youtube,bertasius2020classifying}, self-driving~\cite{caesar2020nuscenes}, and medical imaging~\cite{shen2017deep,litjens2017survey}. Much research has been devoted to explore interactive image segmentation with different interaction types, such as bounding boxes~\cite{xu2017deep}, polygons~\cite{acuna2018efficient}, clicks~\cite{sofiiuk2021reviving}, scribbles~\cite{wu2014milcut}, and their combinations~\cite{zhang2020interactive}. Among them, the click-based approach is most common due to its simplicity and well-established training and evaluation protocols.

Recent advances in click-based approaches mainly lie in two orthogonal directions: 1) the development of more effective backbone networks and 2) the exploration of more elaborate refinement modules built upon the backbone. For the former direction, different hierarchical backbones, including both ConvNets~\cite{lin2022focuscut,sofiiuk2021reviving} and ViTs~\cite{chen2022focalclick,liu2022isegformer}, have been developed for interactive segmentation. For the latter direction, various refinement modules, including local refinement~\cite{chen2022focalclick,lin2022focuscut} and click imitation~\cite{liu2022pseudoclick}, have been proposed to further boost segmentation performance. In this work, we delve into the former direction and focus on exploring a plain backbone for interactive segmentation.

\begin{figure}[t]
    \includegraphics[width=0.47\textwidth, height=6.8cm]{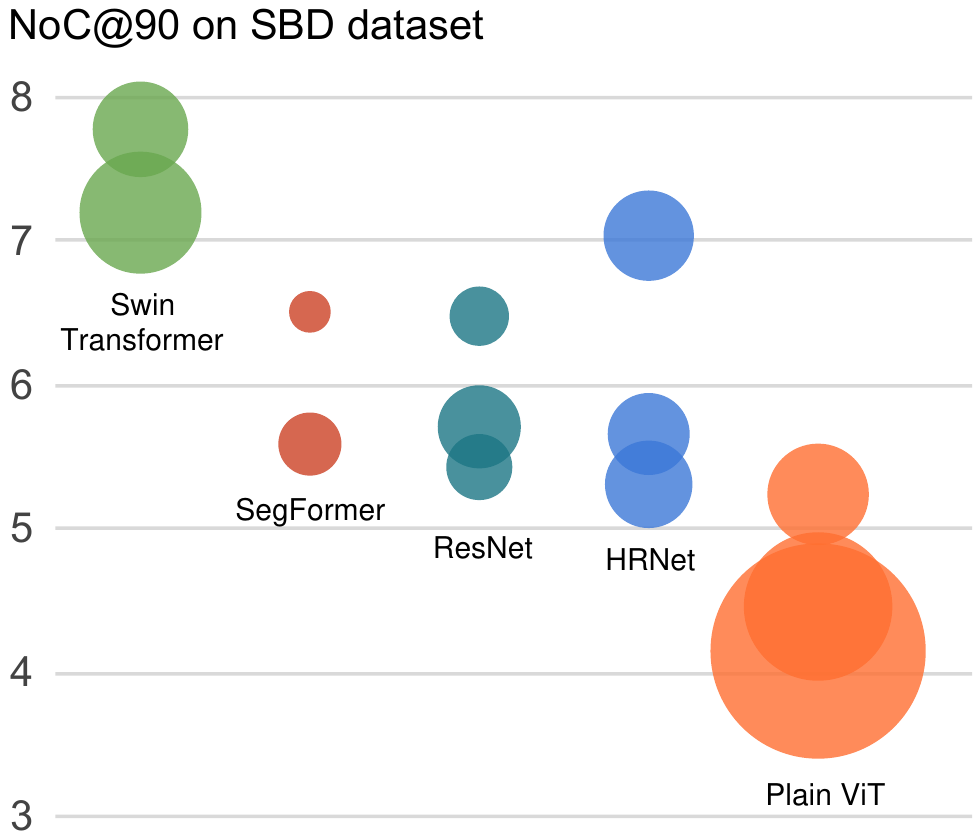}
    \caption{\textbf{Interactive segmentation results on SBD~\cite{hariharan2011semantic}}. The metric ``NoC@90" denotes the number of clicks required to obtain 90\% IoU. The area of each bubble is proportional to the FLOPs of a model variant (Tab.~\ref{tab:computation_analysis}). We show that plain ViTs outperform all hierarchical backbones for interactive image segmentation at a moderate computational cost.}
  \label{fig:teaser}
\end{figure}

A hierarchical backbone is the predominant architecture for current interactive segmentation methods. This design is deeply rooted in ConvNets, represented by ResNet~\cite{he2016deep}, and has been adopted by ViTs, represented by the Swin Transformer~\cite{liu2021swin}. The motivation for a hierarchical backbone stems from the locality of convolution operations, leading to insufficient model receptive field size without the hierarchy. To increase the receptive field size, ConvNets have to progressively downsample feature maps to capture more global contextual information. Therefore, they often require a feature pyramid network such as FPN~\cite{lin2017feature} to aggregate multi-scale representations for high-quality segmentation. However, this reasoning no longer applies for a plain ViT, in which global information can be captured from the first self-attention block. Because all feature maps in the ViT are of the same resolution, the motivation for an FPN-like feature pyramid also no longer remains. The above reasoning is supported by a recent finding that a plain ViT can serve as a strong backbone for object detection~\cite{li2022exploring}. This finding indicates a general-purpose ViT backbone might be suitable for other tasks, which then can decouple pretraining from finetuning and transfer the benefits from readily available pretrained ViT models (\eg MAE~\cite{he2021masked}) to these tasks. However, although this design is simple and has been proven effective, it has not yet been explored in interactive segmentation. In this work, we propose {\SimpleClick}, the first plain-backbone method for interactive segmentation. The core of {\SimpleClick} is a plain ViT backbone that maintains single-scale representations throughout. We \emph{only} use the last feature map from the plain backbone to build a simple feature pyramid for segmentation, largely decoupling the general-purpose backbone from the segmentation-specific modules. To make {\SimpleClick} more efficient, we use a light-weight MLP decoder to transform the simple feature pyramid into a segmentation (see Sec.~\ref{sec:method} for details).

We extensively evaluate our method on \textbf{10} public benchmarks, including both natural and medical images. With the plain backbone pretrained as a MAE~\cite{he2021masked}, our method achieves \textbf{4.15} NoC@90 on SBD, which outperforms the previous best method by \textbf{21.8\%} without a complex FPN-like design and local refinement. We demonstrate the generalizability of our method by out-of-domain evaluation on medical images. We further analyze the computational efficiency of {\SimpleClick}, highlighting its suitability as a practical annotation tool. 

Our main contributions are:
\begin{itemize}[leftmargin=*,noitemsep,topsep=0pt]
    \item We propose {\SimpleClick}, the first plain-backbone method for interactive image segmentation. 
    \item {\SimpleClick} achieves state-of-the-art performance on natural images and shows strong generalizability on medical images.
    \item {\SimpleClick} meets the computational efficiency requirement for a practical annotation tool, highlighting its readiness for real-world applications.
\end{itemize}

\section{Related Work}
\label{sec:related_work}

\noindent\textbf{Interactive Image Segmentation}
Interactive image segmentation is a longstanding problem for which increasingly better solution approaches have been proposed. Early works~\cite{boykov2001interactive,grady2006random,gulshan2010geodesic,rother2004grabcut} tackle this problem using graphs defined over image pixels. However, these methods only focus on low-level image features, and therefore tend to have difficulty with complex objects.

Thriving on large datasets, ConvNets~\cite{xu2016deep,xu2017deep,sofiiuk2021reviving,chen2022focalclick,lin2022focuscut} have evolved as the dominant architecture for high quality interactive segmentation. ConvNet-based methods have explored various interaction types, such as bounding boxes~\cite{xu2017deep}, polygons~\cite{acuna2018efficient}, clicks~\cite{sofiiuk2021reviving}, and scribbles~\cite{wu2014milcut}. Click-based approaches are the most common due to their simplicity and well-established training and evaluation protocols. Xu \etal~\cite{xu2016deep} first proposed a click simulation strategy that has been adopted by follow-up work~\cite{sofiiuk2021reviving,chen2022focalclick,liu2022pseudoclick}.
DEXTR~\cite{maninis2018deep} extracts a target object from specifying its four extreme points (left-most, right-most, top, bottom pixels). FCA-Net~\cite{lin2020interactive} demonstrates the critical role of the first click for better segmentation. Recently, ViTs have been applied to interactive segmentation. FocalClick~\cite{chen2022focalclick} uses SegFormer~\cite{xie2021segformer} as the backbone network and achieves state-of-the-art segmentation results with high computational efficiency. iSegFormer~\cite{liu2022isegformer} uses a Swin Transformer~\cite{liu2021swin} as the backbone network for interactive segmentation on medical images. Besides the contribution on backbones, some works are exploring elaborate refinement modules built upon the backbone. FocalClick~\cite{chen2022focalclick} and FocusCut~\cite{lin2022focuscut} propose similar local refinement modules for high-quality segmentation. PseudoClick~\cite{liu2022pseudoclick} proposes a click-imitation mechanism by estimating the next-click to further reduce human annotation cost. Our method differs from all previous click-based methods in its plain, non-hierarchical ViT backbone, enjoying the benefits from readily available pretrained ViT models (\eg MAE~\cite{he2021masked}).

\noindent\textbf{Vision Transformers for Non-Interactive Segmentation}
Recently, ViT-based approaches~\cite{khan2022transformers,strudel2021segmenter,xie2021segformer,gu2022multi,yuan2021hrformer} have shown competitive performance on segmentation tasks compared to ConvNets. The original ViT~\cite{dosovitskiy2020image} is a non-hierarchical architecture that only maintains single-scale feature maps throughout. SETR~\cite{zheng2021rethinking} and Segmenter~\cite{strudel2021segmenter} use the original ViT as the encoder for semantic segmentation. To allow for more efficient segmentation, the Swin Transformer~\cite{liu2021swin} reintroduces a computational hierarchy into the original ViT architecture using shifted window attention, leading to a highly efficient hierarchical ViT backbone. SegFormer~\cite{xie2021segformer} designs hierarchical feature representations based on the original ViT using overlapped patch merging, combined with a light-weight MLP decoder for efficient segmentation. HRViT~\cite{gu2022multi} integrates a high-resolution multi-branch architecture with ViTs to learn multi-scale representations. Recently, the original ViT has been reintroduced as a competitive backbone for semantic segmentation~\cite{chen2021simple} and object detection~\cite{li2022exploring}, with the aid of MAE~\cite{he2021masked} pretraining and window attention. Inspired by this finding, we explore using a plain ViT as the backbone network for interactive segmentation.

\begin{figure*}[t]
    \centering
    \includegraphics[width=13.0cm, height=7.0cm]{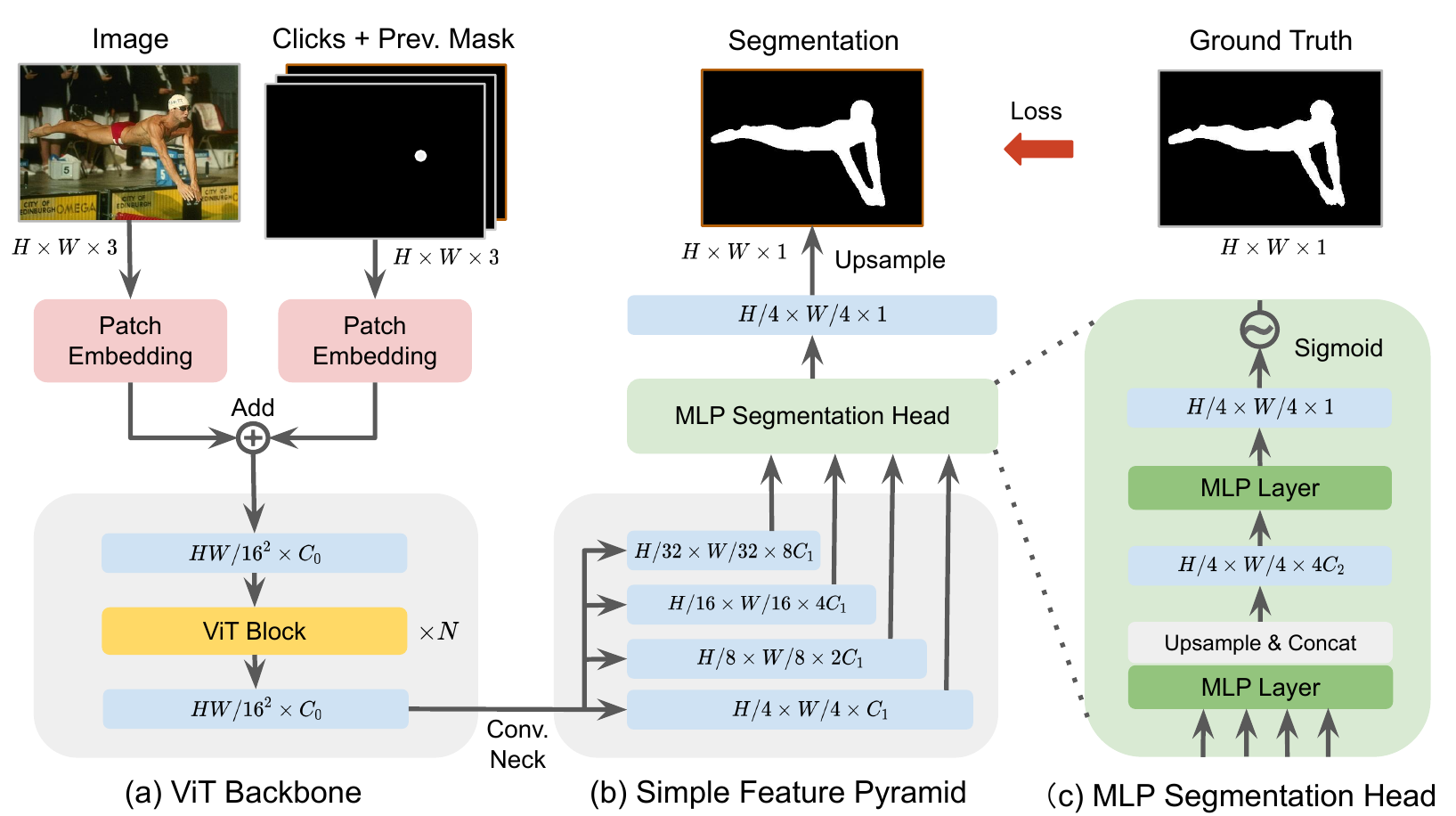}
   \caption{\textbf{{\SimpleClick} overview}. Our method consists of three main modules: (a) a plain ViT backbone that maintains single-scale feature maps throughout; (b) a multi-scale simple feature pyramid that is generated from the \emph{last} feature map of the backbone by four parallel convolution or deconvolution layers; (c) a light-weight MLP decoder for segmentation. We also input the previous segmentation to improve the performance and to allow clicking from existing masks. User clicks are encoded as a two-channel disk map, combined with the previous segmentation as input. Boxes in blue are intermediate feature maps. The positional encoding is not shown for brevity.}
   \label{fig:simpleclick_framework}
\end{figure*}

\section{Method}
\label{sec:method}

Our goal is not to propose new modules, but to adapt a plain-ViT backbone for interactive segmentation with \emph{minimal} modifications so as to enjoy the readily available pretrained ViT weights. Sec.~\ref{sec:method_network_architecture} introduces the main modules of SimpleClick. Sec.~\ref{sec:method_training_inference} describes the training and inference details of our method. 

\subsection{Network Architecture}
\label{sec:method_network_architecture}

\noindent\textbf{Adaptation of Plain-ViT Backbone}
We use a plain ViT~\cite{dosovitskiy2020image} as our backbone network, which only maintains single-scale feature maps throughout. The patch embedding layer divides the input image into non-overlapping fixed-size patches (\eg 16$\times$16 for ViT-B), each patch is flattened and linearly projected to a fixed-length vector (\eg 768 for ViT-B). The resulting sequence of vectors is fed into a queue of Transformer blocks (\eg. 12 for ViT-B) for self-attention. We implement {\SimpleClick} with three backbones: ViT-B, ViT-L, and ViT-H (Tab.~\ref{tab:model_size_comparison} shows the number of parameters for the three backbones). The three backbones were pretrained on ImageNet-1k as MAEs~\cite{he2021masked}. We adapt the pretrained backbones to higher-resolution inputs during finetuning using non-shifting window attention aided by a few global self-attention blocks (\eg 2 for ViT-B), as introduced in ViTDet~\cite{li2022exploring}. Since the last feature map is subject to all the attention blocks, it should have the strongest representation.
Therefore, we only use the last feature map to build a simple multi-scale feature pyramid.

\begin{table}
\footnotesize
\renewcommand\arraystretch{0.9}
  \centering
  \begin{tabular}{l c c c}
    \toprule
    Model$\downarrow$ Module$\rightarrow$ & ViT Backbone & Conv. Neck & MLP Head \\
    \midrule
    Ours-ViT-B & 83.0 (89.3\%) & 9.0 (9.7\%) & 0.9 (1.0\%) \\
    Ours-ViT-L & 290.8 (94.3\%) & 16.5 (5.3\%) & 1.1 (0.4\%)\\
    Ours-ViT-H & 604.0 (95.7\%) & 25.8 (4.1\%) & 1.3 (0.2\%) \\
    \bottomrule
  \end{tabular}
  \caption{\textbf{Number of parameters of our models}. The unit is million. The percentage of parameters is shown in bracket. Most parameters are used by the ViT backbone.} 
  \label{tab:model_size_comparison}
\end{table}

\noindent\textbf{Simple Feature Pyramid}
For the hierarchical backbone, a feature pyramid is commonly produced by an FPN~\cite{lin2017feature} to combine features from different stages. For the plain backbone, a feature pyramid can be generated in a much simpler way: by a set of parallel convolutional or deconvolutional layers using \emph{only} the last feature map of the backbone. As shown in Fig.~\ref{fig:simpleclick_framework}, given the input ViT feature map, a multi-scale feature map can be produced by four convolutions with different strides. Though the effectiveness of this simple feature pyramid design is first demonstrated in ViTDet~\cite{li2022exploring} for object detection, we show in this work the effectiveness of this simple feature pyramid design for interactive segmentation. We also propose several additional variants (Fig.~\ref{fig:ablation_feature_pyramid}) as part of an ablation study (Sec.~\ref{sec:ablation_study}).

\noindent\textbf{All-MLP Segmentation Head}
We implement a lightweight segmentation head using only MLP layers. It takes in the simple feature pyramid and produces a segmentation probability map\footnote{This probability map may be miscalibrated and can be improved by calibration approaches~\cite{ding2021local}.} of scale $1/4$, followed by an upsampling operation to recover the original resolution. Note that this segmentation head avoids computationally demanding components and only accounts for up to 1\% of the model parameters (Tab.~\ref{tab:model_size_comparison}). The key insight is that with a powerful pretrained backbone, a lightweight segmentation head is sufficient for interactive segmentation. The proposed all-MLP segmentation head works in three steps. \emph{First}, each feature map from the simple feature pyramid goes through an MLP layer to transform it to an identical channel dimension (\ie $C_2$ in Fig.~\ref{fig:simpleclick_framework}). \emph{Second}, all feature maps are upsampled to the same resolution (\ie $1/4$ in Fig.~\ref{fig:simpleclick_framework}) for concatenation. \emph{Third}, the concatenated features are fused by another MLP layer to produce a single-channel feature map, followed by a sigmoid function to obtain a segmentation probability map, which is then transformed to a binary segmentation given a predefined threshold (\ie 0.5).

\noindent\textbf{Symmetric Patch Embedding and Beyond}
To fuse human clicks into the plain backbone, we introduce a patch embedding layer that is symmetric to the patch embedding layer in the backbone, followed by element-wise feature addition. The user clicks are encoded in a two-channel disk map, one for positive clicks and the other for negative clicks. The positive clicks should be placed on the foreground, while the negative clicks should be placed on the background. The previous segmentation and the two-channel click map are concatenated as a three-channel map for patch embedding. The two symmetric embedding layers operate on the image and the concatenated three-channel map, respectively. The inputs are patchified, flattened, and projected to two vector sequences of the same dimension, followed by element-wise addition before inputting into the self-attention blocks.

\subsection{Training and Inference Settings}
\label{sec:method_training_inference}

\noindent\textbf{Backbone Pretraining} 
Our backbone models are pretrained as MAEs~\cite{he2021masked} on ImageNet-1K~\cite{deng2009imagenet}. In MAE pretraining, the ViT models reconstruct the randomly masked pixels of images while learning a universal representation. This simple self-supervised approach turns out to be an efficient and scalable way to pretrain ViT models~\cite{he2021masked}. In this work, we do not perform pretraining ourselves. Instead, we simply use the readily available pretrained MAE weights from~\cite{he2021masked}.

\noindent\textbf{End-to-end Finetuning}
With the pretrained backbone, we finetune our model end-to-end on the interactive segmentation task. The finetuning pipeline can be briefly described as follows. \emph{First}, we automatically simulate clicks based on the current segmentation and gold standard segmentation, without a human-in-the-loop providing the clicks. Specifically, we use a combination of random and iterative click simulation strategies, inspired by RITM~\cite{sofiiuk2021reviving}. The random click simulation strategy generates clicks in parallel, without considering the order of the clicks. The iterative click simulation strategy generates clicks iteratively, where the next click should be placed on the erroneous region of a prediction that was obtained using the previous clicks. This strategy is more similar to human clicking behavior. \emph{Second}, we incorporate the segmentation from the previous interaction as an additional input for the backbone, further improving the segmentation quality. This also allows our method to refine from an existing segmentation, which is a desired feature for a practical annotation tool.
We use the normalized focal loss~\cite{sofiiuk2021reviving} (NFL) to train all our models. Previous works~\cite{sofiiuk2021reviving,chen2022focalclick} show that NFL converges faster and achieves better performance than the widely used binary cross entropy loss for interactive segmentation tasks. 
Similar training pipelines have been proposed by RITM~\cite{sofiiuk2021reviving} and its follow-up works~\cite{chen2022focalclick,liu2022pseudoclick,chen2021conditional}.

\noindent\textbf{Inference}
There are two inference modes: automatic evaluation and human evaluation.
For automatic evaluation, clicks are automatically simulated based on the current segmentation and gold standard. For human evaluation, a human-in-the-loop provides all clicks based on their subjective evaluation of current segmentation results. We use automatic evaluation for quantitative analyses and human evaluation for a qualitative assessment of the interactive segmentation behavior.

\begin{figure*}[t!]
        \includegraphics[width=8.5cm, height=5.8cm, trim=40 5 40 5, clip]{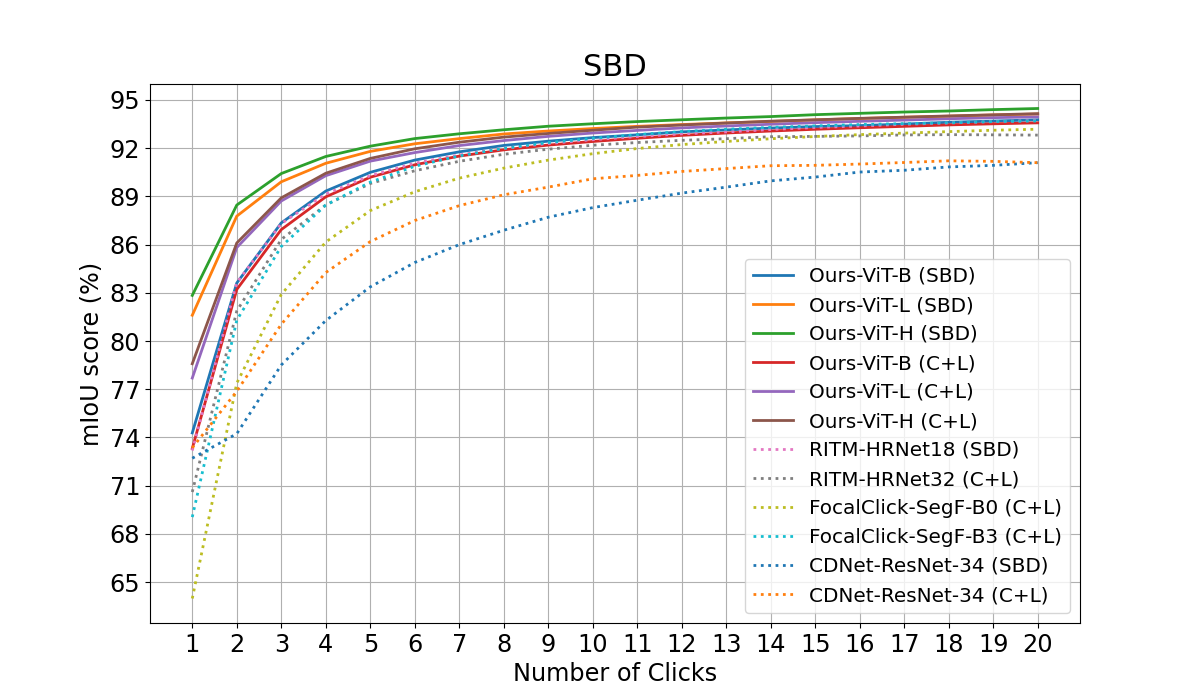}
        \includegraphics[width=8.5cm, height=5.8cm, trim=40 5 40 5, clip]{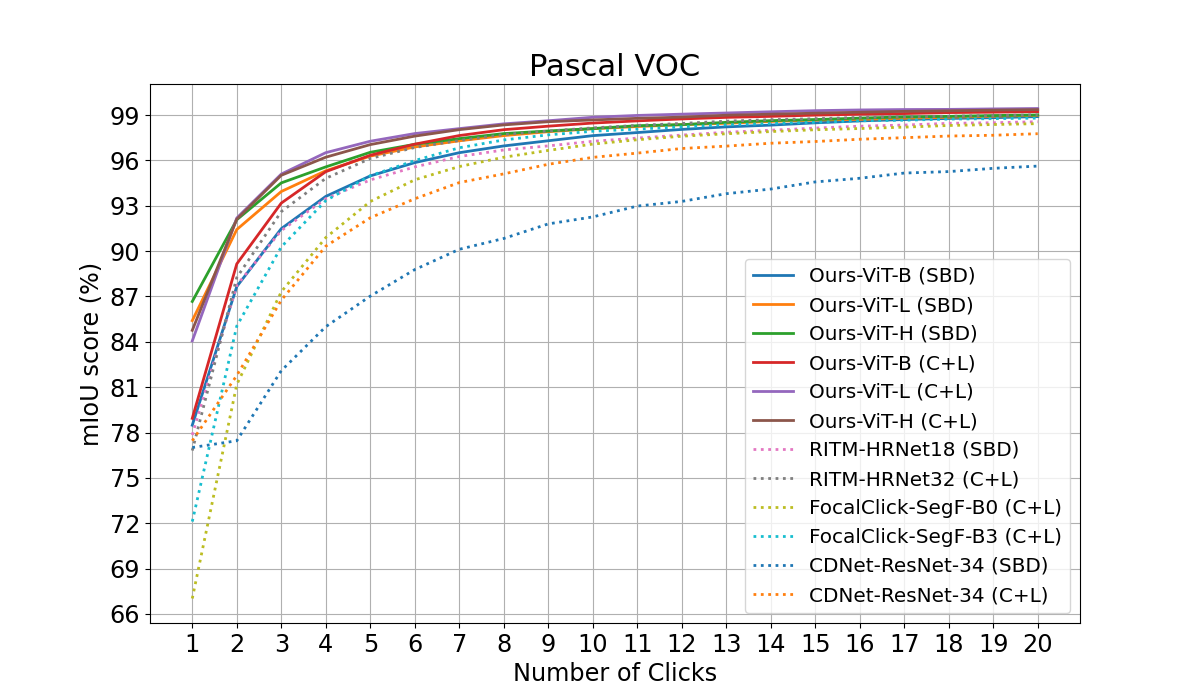}
    \caption{\textbf{Convergence analysis} for models trained on either SBD~\cite{hariharan2011semantic} or COCO~\cite{lin2014microsoft}+LVIS~\cite{gupta2019lvis} (C+L). We report results on SBD~\cite{hariharan2011semantic} and Pascal VOC~\cite{everingham2010pascal}. The metric is mean IoU given $k$ clicks (mIoU@$k$). Our models in general require fewer clicks for a given accuracy level.}
    \label{fig:convergence_analysis}
\end{figure*}

\begin{table*}
  \footnotesize
  \centering
  \begin{tabularx}{\textwidth}{l l c c c c c c c c c c}
    \toprule
    \multirow{2}{*}{Method} & \multirow{2}{*}{Backbone} & 
    \multicolumn{2}{c}{GrabCut} & \multicolumn{2}{c}{Berkeley} & \multicolumn{2}{c}{SBD} & \multicolumn{2}{c}{DAVIS} & \multicolumn{2}{c}{Pascal VOC} \\
    & & NoC85 & NoC90 & NoC85 & NoC90 & NoC85 & NoC90 & NoC85 & NoC90 & NoC85 & NoC90 \\
    \midrule
    $\eighthnote$ DIOS \cite{xu2016deep} \tiny{$_\text{CVPR16}$} & FCN
    & - & 6.04 & - & 8.65 & - & - & - & 12.58 & 6.88 & - \\
    $\eighthnote$ FCA-Net \cite{lin2020interactive} \tiny{$_\text{CVPR20}$} & ResNet-101
    & - & 2.08 & - & 3.92 & - & - & - & 7.57 & 2.69 & - \\
    \midrule
    $\quarternote$ LD \cite{li2018interactive} \tiny{$_\text{CVPR18}$} & VGG-19
    & 3.20 & 4.79 & - & - & 7.41 & 10.78 & 5.05 & 9.57 & - & - \\
    $\quarternote$ BRS  \cite{jang2019interactive} \tiny{$_\text{CVPR19}$} & DenseNet
    & 2.60 & 3.60 & - & 5.08 & 6.59 & 9.78 & 5.58 & 8.24 & - & - \\
    $\quarternote$ f-BRS  \cite{sofiiuk2020f} \tiny{$_\text{CVPR20}$} & ResNet-101
    & 2.30 & 2.72 & - & 4.57 & 4.81 & 7.73 & 5.04 & 7.41 & - & - \\
    $\quarternote$ RITM~\cite{sofiiuk2021reviving} \tiny{$_\text{Preprint21}$} & HRNet-18
    & 1.76 & 2.04 & 1.87 & 3.22 & 3.39 & 5.43 & 4.94 & 6.71 & 2.51 & 3.03 \\  
    $\quarternote$ CDNet~\cite{chen2021conditional} \tiny{$_\text{ICCV21}$} & ResNet-34
    & 1.86 & 2.18 & 1.95 & 3.27 & 5.18 & 7.89 & 5.00 & 6.89 & 3.61 & 4.51 \\ 
    $\quarternote$ PseudoClick~\cite{liu2022pseudoclick} \tiny{$_\text{ECCV22}$} & HRNet-18
    & 1.68 & 2.04 & 1.85 & 3.23 & 3.38 & 5.40 & 4.81 & 6.57 & 2.34 & 2.74 \\
    $\quarternote$ FocalClick~\cite{chen2022focalclick} \tiny{$_\text{CVPR22}$} & HRNet-18s 
    & 1.86 & 2.06 & - & 3.14 & 4.30 & 6.52 & 4.92 & 6.48 & - & - \\
    $\quarternote$ FocalClick~\cite{chen2022focalclick} \tiny{$_\text{CVPR22}$} & SegF-B0 
    & 1.66 & 1.90 & - & 3.14 & 4.34 & 6.51 & 5.02 & 7.06 & - & - \\
    $\quarternote$ FocusCut~\cite{lin2022focuscut} \tiny{$_\text{CVPR22}$} & ResNet-50 
    & 1.60 & 1.78 & 1.85$^\dagger$ & 3.44 & 3.62 & 5.66 & 5.00 & 6.38 & - & - \\
    $\quarternote$ FocusCut~\cite{lin2022focuscut} \tiny{$_\text{CVPR22}$} & ResNet-101
    & 1.46 & 1.64 & 1.81$^\dagger$ & 3.01 & 3.40 & 5.31 & 4.85 & 6.22 & - & - \\
    \rowcolor[gray]{0.9}    
    $\quarternote$ Ours & ViT-B 
    & 1.40 & 1.54 & 1.44 & 2.46 & 3.28 & 5.24 & \textbf{4.10} & 5.48 & 2.38 & 2.81 \\
    \rowcolor[gray]{0.9}
    $\quarternote$ Ours & ViT-L 
    & 1.38 & 1.46 & 1.40 & 2.33 & 2.69 & 4.46 & 4.12 & 5.39 & 1.95 & 2.30 \\
    \rowcolor[gray]{0.9}
    $\quarternote$ Ours & ViT-H
    & \textbf{1.32} & \textbf{1.44} & \textbf{1.36} & \textbf{2.09} & \textbf{2.51} & \textbf{4.15} & 4.20 & \textbf{5.34} & \textbf{1.88} & \textbf{2.20} \\
    \midrule
    $\twonotes$ RITM~\cite{sofiiuk2021reviving} \tiny{$_\text{Preprint21}$} & HRNet-32
    & 1.46 & 1.56 & 1.43 & 2.10 & 3.59 & 5.71 & 4.11 & 5.34 & 2.19 & 2.57 \\    
    $\twonotes$ CDNet~\cite{chen2021conditional} \tiny{$_\text{ICCV21}$} & ResNet-34
    & 1.40 & 1.52 & 1.47 & 2.06 & 4.30 & 7.04 & 4.27 & 5.56 & 2.74 & 3.30 \\ 
    $\twonotes$ PseudoClick~\cite{liu2022pseudoclick} \tiny{$_\text{ECCV22}$} & HRNet-32
    & 1.36 & 1.50 & 1.40 & 2.08 & 3.46 & 5.54 & 3.79 & 5.11 & 1.94 & 2.25 \\
    $\twonotes$ FocalClick~\cite{chen2022focalclick} \tiny{$_\text{CVPR22}$} & SegF-B0
    & 1.40 & 1.66 & 1.59 & 2.27 & 4.56 & 6.86 & 4.04 & 5.49 & 2.97 & 3.52 \\
    $\twonotes$ FocalClick~\cite{chen2022focalclick} \tiny{$_\text{CVPR22}$} & SegF-B3
    & 1.44 & 1.50 & 1.55 & 1.92 & 3.53 & 5.59 & 3.61 & 4.90 & 2.46 & 2.88 \\
    \rowcolor[gray]{0.9}    
    $\twonotes$ Ours & ViT-B
    & 1.38 & 1.48 & 1.36 & 1.97 & 3.43 & 5.62 & 3.66 & 5.06 & 2.06 & 2.38 \\
    \rowcolor[gray]{0.9}
    $\twonotes$ Ours & ViT-L
    & \textbf{1.32} & \textbf{1.40} & \textbf{1.34} & 1.89 & 2.95 & 4.89 & \textbf{3.26} & 4.81 & \textbf{1.72} & \textbf{1.96}  \\
    \rowcolor[gray]{0.9}
    $\twonotes$ Ours & ViT-H
    & 1.38 & 1.50 & 1.36 & \textbf{1.75} & \textbf{2.85} & \textbf{4.70} & 3.41 & \textbf{4.78} & 1.76 & 1.98 \\
    \bottomrule
  \end{tabularx}
  \caption{\textbf{Comparison with previous results.} We report results on five benchmarks: GrabCut~\cite{rother2004grabcut}, Berkeley~\cite{martin2001database}, SBD~\cite{hariharan2011semantic}, DAVIS~\cite{perazzi2016benchmark}, and Pascal VOC~\cite{everingham2010pascal}. The best results are set in bold. $\eighthnote$ denotes a model trained on Pascal; $\quarternote$ denotes a model trained on SBD; $\twonotes$ denotes a model trained on COCO~\cite{lin2014microsoft}+LVIS~\cite{gupta2019lvis} (C+L); $\dagger$ denotes a number reproduced by the released or retrained models. Our models achieve state-of-the-art performance on all benchmarks.}
  \label{tab:comparision_sota_noc}
\end{table*}

\section{Experiments}
\label{sec:experiments}

\noindent\textbf{Datasets} 
We conducted experiments on \textbf{10} public datasets including 7 natural image datasets and 3 medical datasets.
The details are as follows:
\begin{itemize}[leftmargin=*,noitemsep,topsep=0pt]
    \item \textbf{GrabCut}~\cite{rother2004grabcut}: 50 images (50 instances), each with clear foreground and background differences.
    \item \textbf{Berkeley}~\cite{martin2001database}: 96 images (100 instances); this dataset shares a small portion of images with GrabCut.
    \item \textbf{DAVIS}~\cite{perazzi2016benchmark}: 50 videos; we only use the same 345 frames as used in~\cite{sofiiuk2021reviving,lin2022focuscut,chen2022focalclick,liu2022pseudoclick} for evaluation.
    \item \textbf{Pascal VOC}~\cite{everingham2010pascal}: 1449 images (3427 instances) in the validation set. We only test on the validation set.  
    \item \textbf{SBD}~\cite{hariharan2011semantic}: 8498 training images (20172 instances) and 2857 validation images (6671 instances). Following previous works~\cite{sofiiuk2021reviving,chen2022focalclick,lin2022focuscut}, we train our model on the training set and evaluate on the validation set.
    \item \textbf{COCO}~\cite{lin2014microsoft}+\textbf{LVIS}~\cite{gupta2019lvis} (C+L): COCO contains 118K training images (1.2M instances); LVIS shares the same images with COCO but has much higher segmentation quality. We combine the two datasets for training.
    \item \textbf{ssTEM}~\cite{gerhard2013segmented}: two image stacks, each contains 20 medical images. We use the same stack that was used in~\cite{liu2022pseudoclick}.
    \item \textbf{BraTS}~\cite{baid2021rsna}: 369 magnetic resonance image (MRI) volumes; we test on the same 369 slices used in~\cite{liu2022pseudoclick}.
    \item \textbf{OAIZIB}~\cite{ambellan2019automated}: 507 MRI volumes; we test on the same 150 slices (300 instances) as used in~\cite{liu2022isegformer}.
\end{itemize}

\noindent\textbf{Evaluation Metrics}
Following previous works~\cite{sofiiuk2020f,sofiiuk2021reviving,lin2022focuscut}, we automatically simulate user clicks by comparing the current segmentation with the gold standard. In this simulation, the next click will be put at the center of the region with the largest error. We use the Number of Clicks (NoC) as the evaluation metric to calculate the number of clicks required to achieve a target Intersection over Union (IoU). We set two target IoUs: 85\% and 90\%, represented by NoC\%85 and NoC\%90 respectively. The maximum number of clicks for each instance is set to 20. We also use the average IoU given $k$ clicks (mIoU@$k$) as an evaluation metric to measure the segmentation quality given a fixed number of clicks. 

\begin{figure*}[t!]
    \includegraphics[width=6.0cm, height=4.0cm]{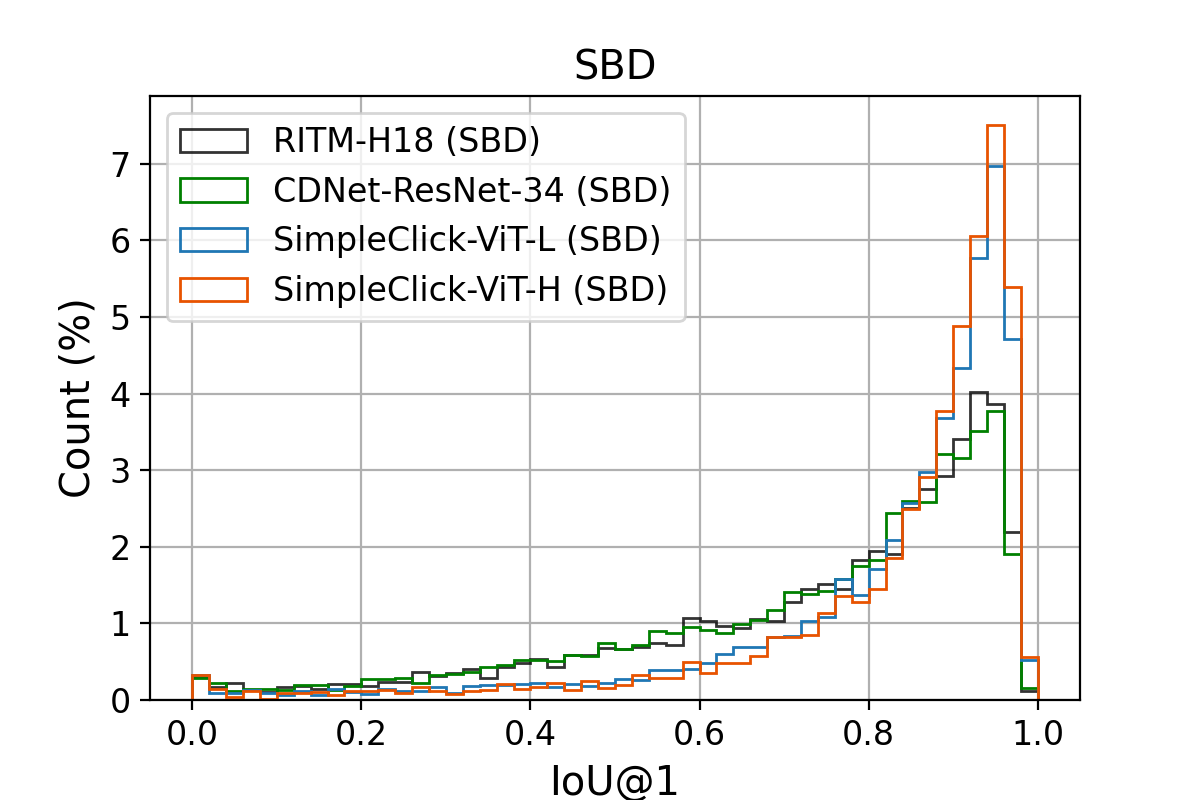}
    \includegraphics[width=6.0cm, height=4.0cm]{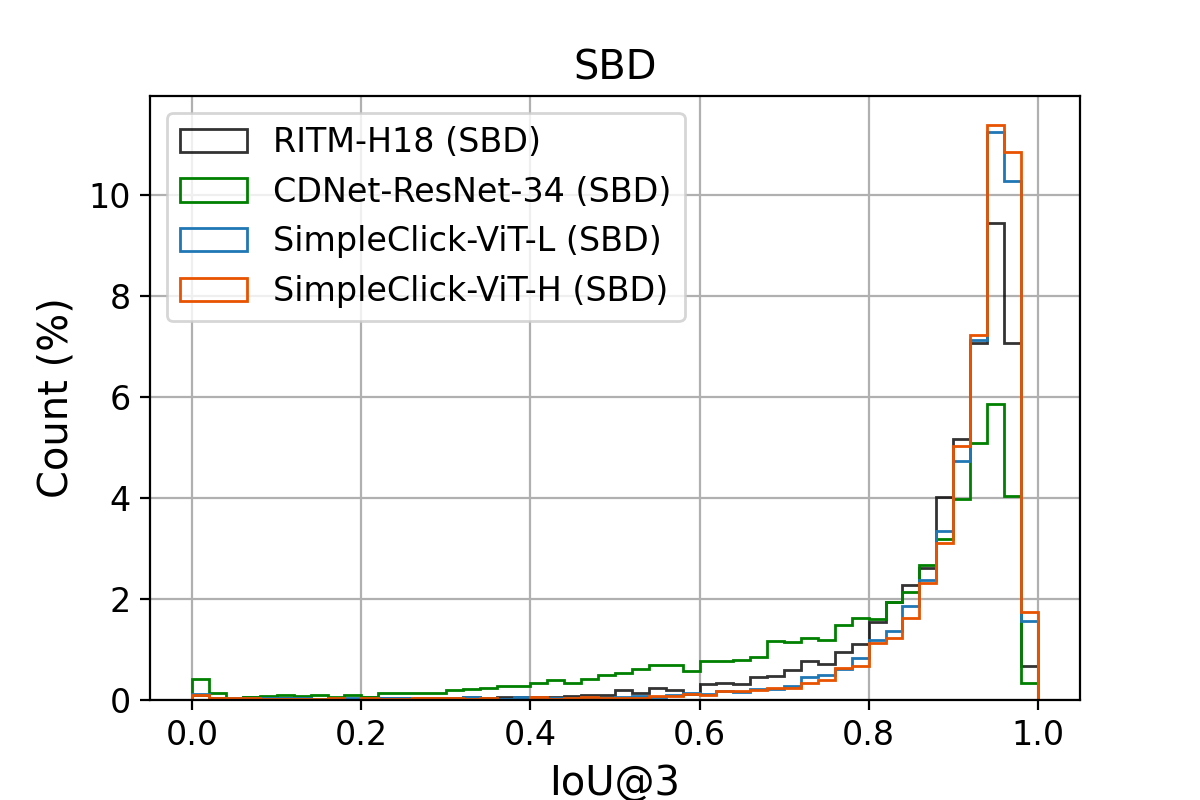}
    \includegraphics[width=6.0cm, height=4.0cm]{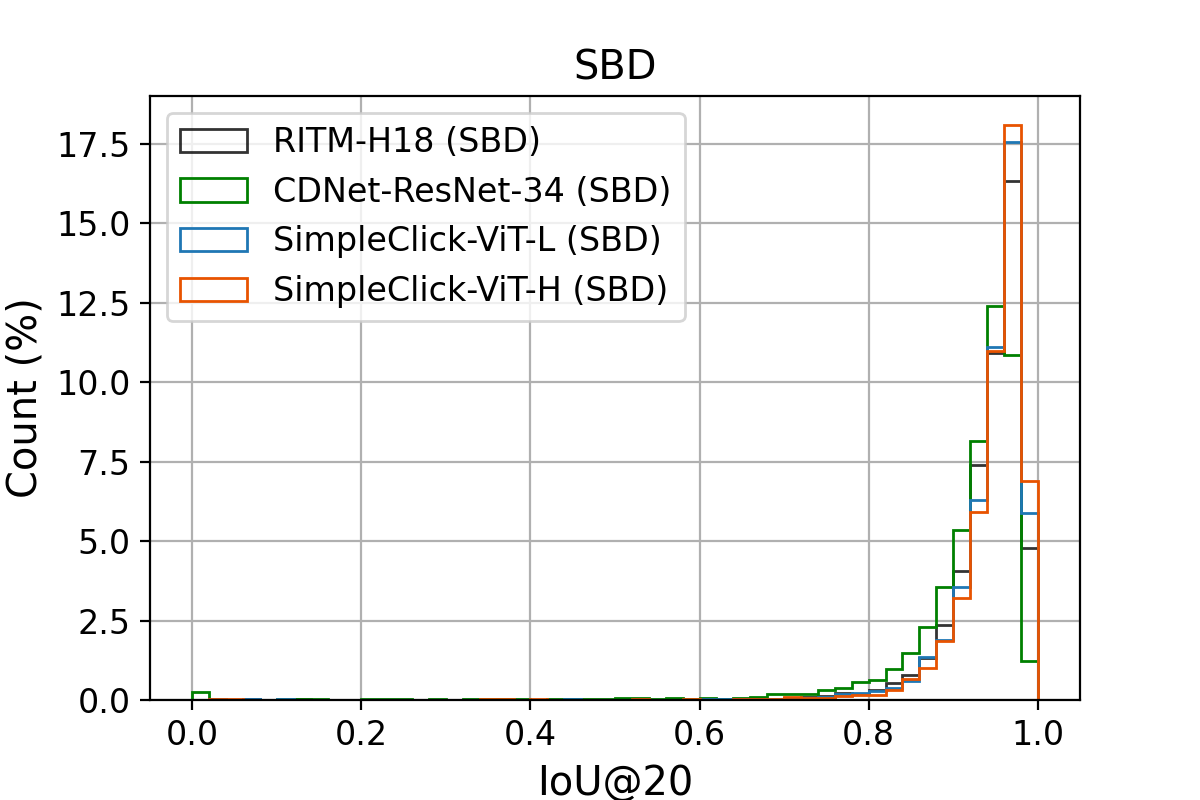}
    \includegraphics[width=6.0cm, height=4.0cm]{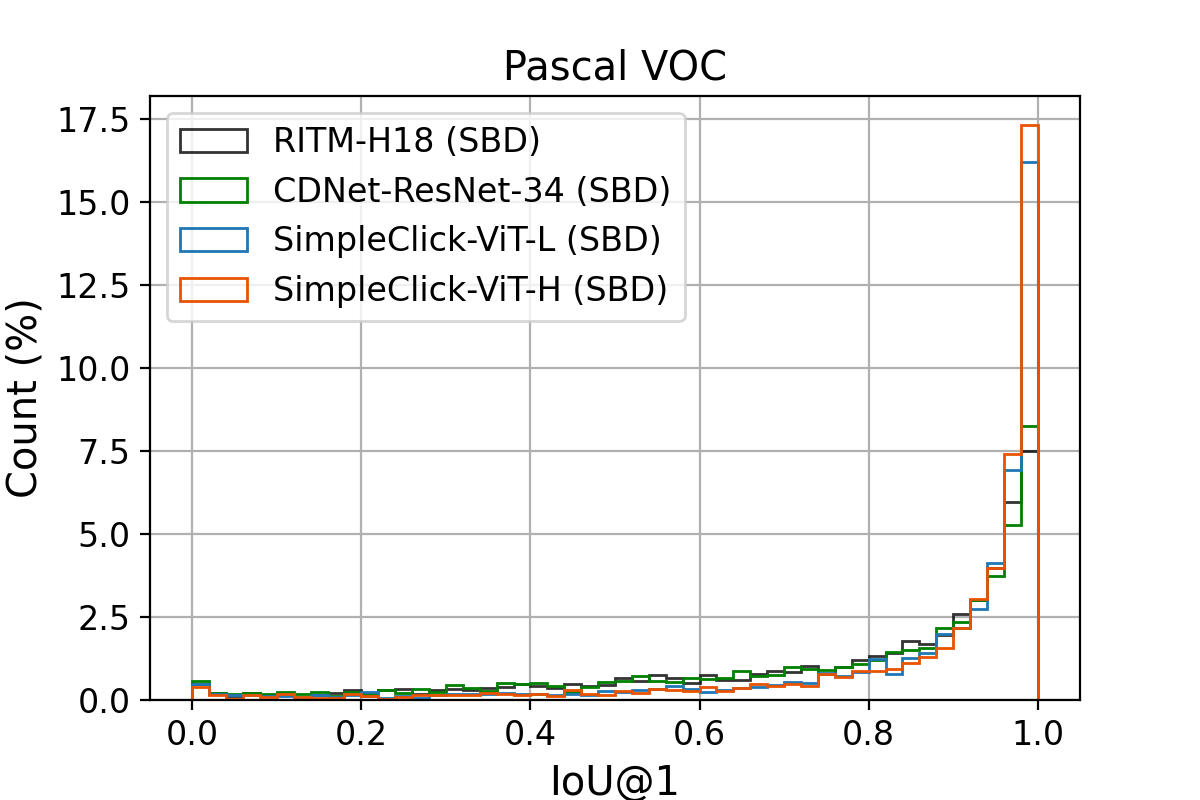}
    \includegraphics[width=6.0cm, height=4.0cm]{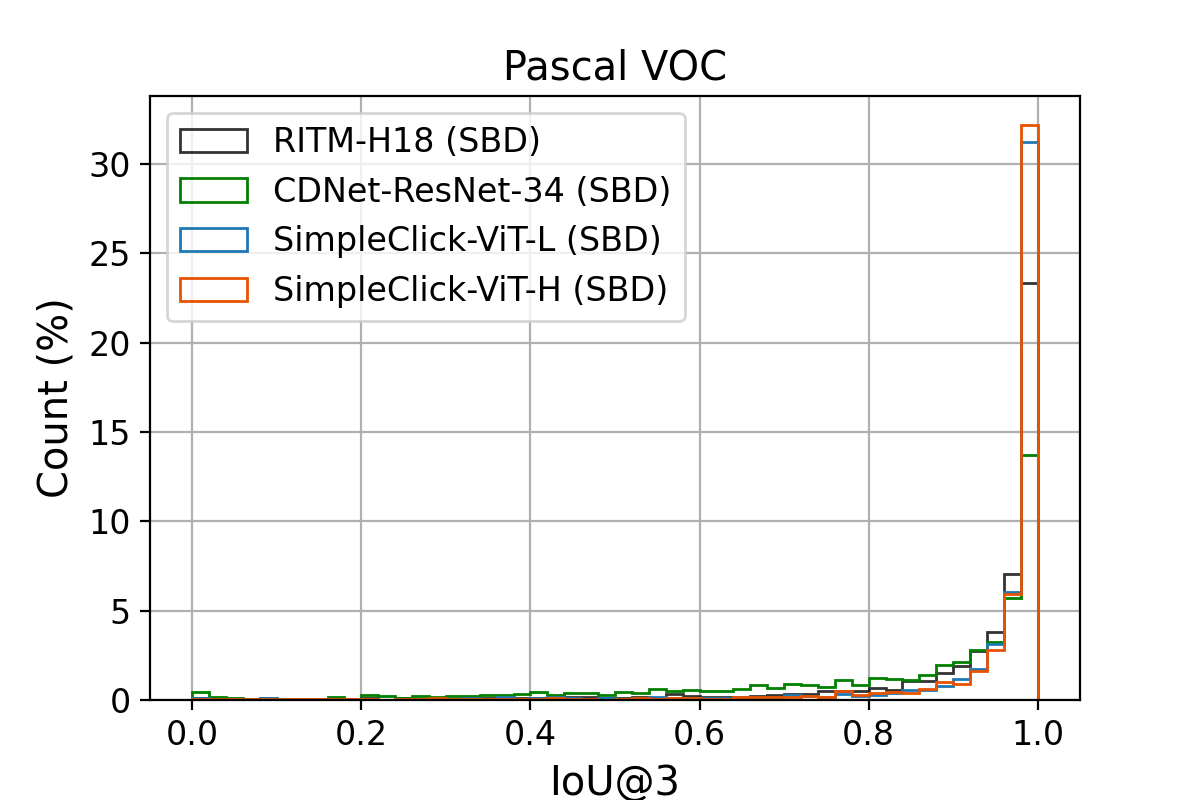}
    \includegraphics[width=6.0cm, height=4.0cm]{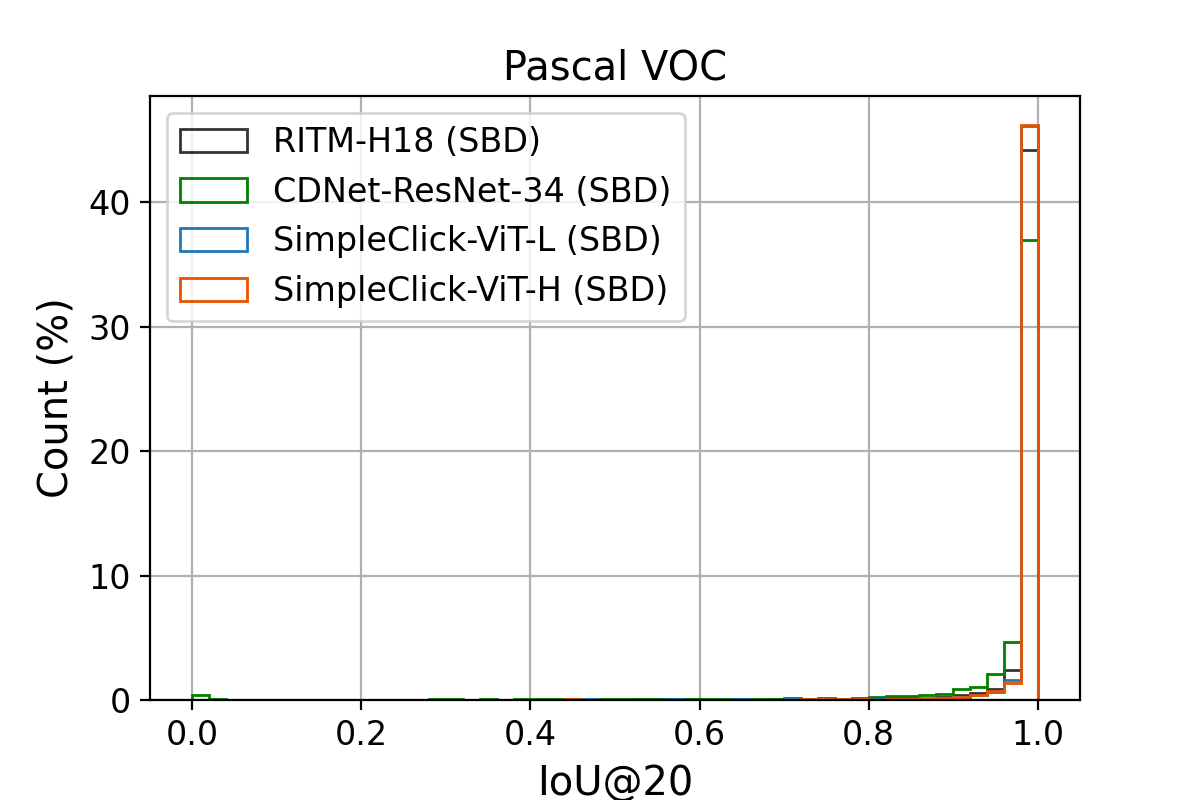}
    \caption{\textbf{Histogram analysis} of IoU given a predefined number of clicks $k$ (IoU@k). We report analysis on SBD~\cite{hariharan2011semantic} and Pascal VOC~\cite{everingham2010pascal} with models trained on SBD. Compared with the two baselines, our models achieve higher-quality segmentation with fewer failure cases.}
    \label{fig:histogram}
\end{figure*}

\begin{figure*}[t!]
    \includegraphics[width=8.5cm, height=5.8cm, trim=40 5 40 5, clip]{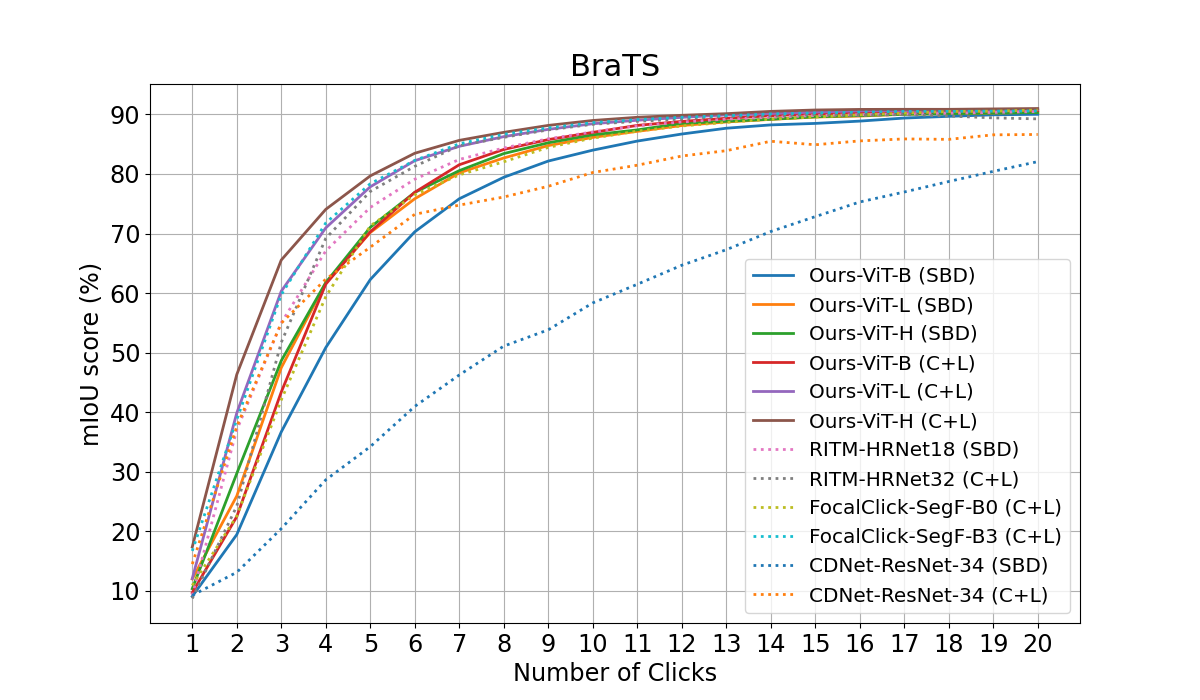}
    \includegraphics[width=8.5cm, height=5.8cm, trim=40 5 40 5, clip]{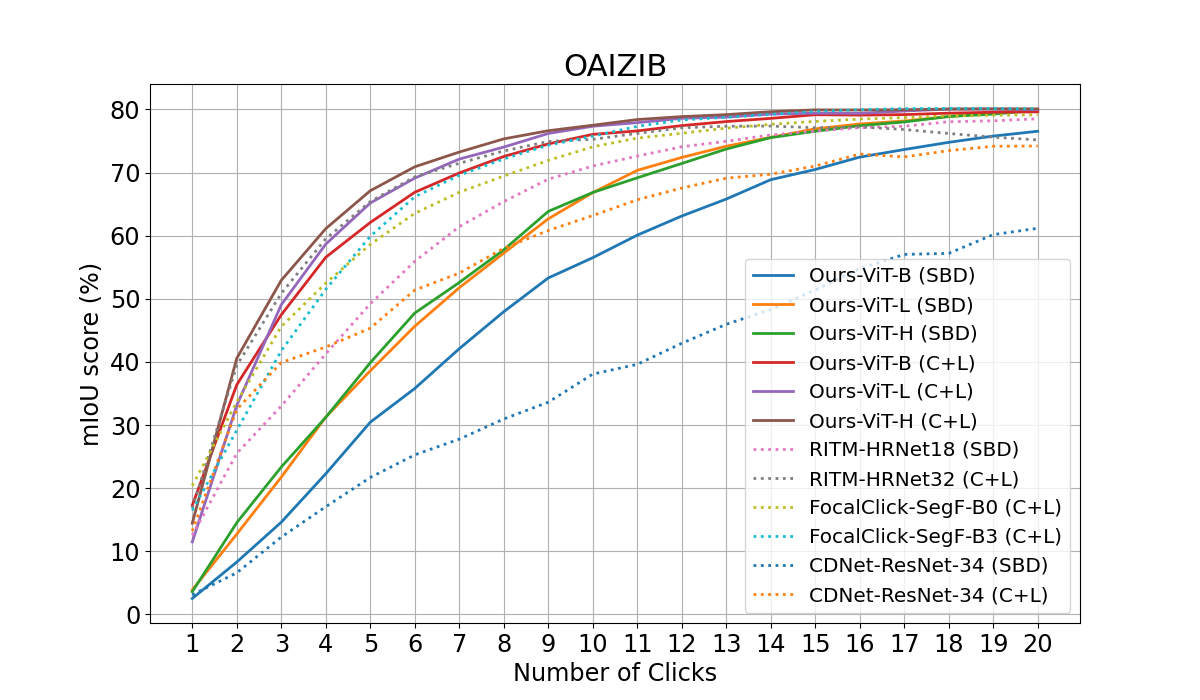}
    \caption{\textbf{Convergence analysis} for two medical image datasets: BraTS~\cite{baid2021rsna} and OAIZIB~\cite{ambellan2019automated}. Models are trained on either SBD~\cite{hariharan2011semantic} or COCO~\cite{lin2014microsoft}+LVIS~\cite{gupta2019lvis} (denoted as C+L). The metric is mean IoU given $k$ clicks. Overall, our models require fewer clicks for a given accuracy level. The performance gain is more prominent for bigger models (\eg ViT-H) or larger training sets (\eg C+L).}
    \label{fig:convergence_analysis_medical}
\end{figure*}

\noindent\textbf{Implementation Details}
We implement our models using Python and PyTorch~\cite{paszke2019pytorch}. We implement three models based on three vanilla ViT models (\ie ViT-B, ViT-L, and ViT-H). These backbone models are initialized with the MAE pretrained weights, and then are finetuned end-to-end with other modules. We train our models on either SBD or COCO+LVIS with 55 epochs; the initial learning rate is set to $5\times10^{-5}$ and decreases to $5\times10^{-6}$ after epoch 50. We set the batch size to 140 for ViT-Base, 72 for ViT-Large, and 32 for ViT-Huge to fit the models into GPU memory. All our models are trained on four NVIDIA RTX A6000 GPUs. We use the following data augmentation techniques: random resizing (scale range from 0.75 to 1.25), random flipping and rotation, random brightness contrast, and random cropping. Though the ViT backbone was pretrained on images of size 224$\times$224, we finetune on $448\times448$ with non-shifting window attention for better performance. We optimize using Adam with $\beta_1=0.9$, $\beta_2=0.999$. 

\subsection{Comparison with Previous Results}
\label{sec:comparison-with-previous-results}

We show in Tab.~\ref{tab:comparision_sota_noc} the comparisons with previous state-of-the-art results. Our models achieves the best performance on all the five benchmarks. Remarkably, when trained on SBD training set, our ViT-H model achieves 4.15 NoC@90 on the SBD validation set, outperforming the previous best score by 21.8\%. Since the SBD validation set contains the largest number of instances (6671 instances) among the five benchmarks, this improvement is convincing. When trained on COCO+LVIS, our models also achieve state-of-the-art performance on all benchmarks. Fig.~\ref{fig:qualitative-results} shows several segmentation cases on DAVIS, including the worst case. Note that the DAVIS dataset requires high-quality segmentations because all its instances have a high-quality gold standard. Our models still achieve the state-of-the-art on DAVIS without using specific modules, such as a local refinement module~\cite{chen2022focalclick}, which is beneficial for high-quality segmentation. Fig.~\ref{fig:convergence_analysis} shows that our method converges better than other methods with sufficient clicks, leading to fewer failure cases as shown in Fig.~\ref{fig:histogram}. We only report results on SBD and Pascal VOC, the top two largest datasets.

\subsection{Out-of-Domain Evaluation on Medical Images}
\label{sec:OOD-evaluation-on-medical-images}

We further evaluate the generalizability of our models on three medical image datasets: ssTEM~\cite{gerhard2013segmented}, BraTS~\cite{bai2007geodesic}, and OAIZIB~\cite{ambellan2019automated}. Tab.~\ref{tab:ood_evaluation_on_medical_images} reports the evaluation results on these three datasets. Fig.~\ref{fig:convergence_analysis_medical} shows the convergence analysis on BraTS and OAIZIB. Overall, our models generalize well to medical images. We also find that the models trained on larger datasets (\ie C+L) generalize better than the models trained on smaller datasets (\ie SBD). 

\begin{table}
\footnotesize
\centering
\begin{tabular}{l c c c}
    \toprule
    \multirow{2}{*}{Model} & ssTEM & BraTS & OAIZIB \\
    & mIoU@10 & mIoU@10 / 20 & mIoU@10 / 20 \\
    \midrule
    $\quarternote$ RITM-H18~\cite{sofiiuk2021reviving}             & 93.15  & 87.05 / 90.47 & 71.04 / 78.52 \\
    $\quarternote$ CDN-RN34~\cite{chen2021conditional}          & 66.72  & 58.34 / 82.07 & 38.07 / 61.17 \\
    $\twonotes$ RITM-H32~\cite{sofiiuk2021reviving}                & 94.11  & 88.34 / 89.25 & 75.27 / 75.18 \\
    $\twonotes$ CDN-RN34~\cite{chen2021conditional}             & 88.46  & 80.24 / 86.63 & 63.19 / 74.21 \\
    $\twonotes$ FC-SF-B0~\cite{chen2022focalclick}           & 92.62  & 86.02 / 90.74 & 74.08 / 79.14 \\
    $\twonotes$ FC-SF-B3~\cite{chen2022focalclick}           & 93.61  & 88.62 / 90.58 & 75.77 / 80.08 \\
    \rowcolor[gray]{0.9}    
    $\twonotes$ Ours-ViT-B              & 93.72  & 86.98 / 90.67 & 76.05 / 79.61\\
    \rowcolor[gray]{0.9}    
    $\twonotes$ Ours-ViT-L              & \textbf{94.34}  & 88.43 / 90.84 & 77.34 / 79.97 \\
    \rowcolor[gray]{0.9}    
    $\twonotes$ Ours-ViT-H              & 94.08  & \textbf{88.98} / \textbf{91.00} & \textbf{77.50} / \textbf{80.10} \\
    \bottomrule
\end{tabular}
\caption{\textbf{Out-of-domain evaluation} on three medical image datasets: ssTEM~\cite{gerhard2013segmented}, BraTS~\cite{baid2021rsna}, and OAIZIB~\cite{ambellan2019automated}. Our models generalize very well on the three datasets, without finetuning.}
\label{tab:ood_evaluation_on_medical_images}
\end{table}

\subsection{Towards Practical Annotation Tool}
\label{sec:towards-practical-annotation-tool}

\noindent\textbf{Tiny Backbone}  To allow for practical applications, especially on low-end devices with limited computational resources, we implement an extremely tiny backbone (\ie ViT-xTiny) for SimpleClick. Compared with ViT-Base, ViT-xTiny decreases the embedding dimension from 768 to 160 and the number of attention blocks from 12 to 8. We end up with a SimpleClick-xTiny model, which is comparable with the tiny FocalClick models in terms of parameters. Comparison results in Tab.~\ref{tab:tiny_vits} show that our model outperforms FocalClick models, even though it is trained from scratch due to the lack of readily available pretrained weights.

\begin{table}
\footnotesize
\renewcommand\arraystretch{0.9}
\setlength{\tabcolsep}{0.6mm}{
  \centering
  \begin{tabular}{l c c c c c}
    \toprule
    Model & Backbone & Pretrained & Params/M & NoC85 & NoC90\\
    \midrule
    FocalClick  & HRNet-18s-S1    & \cmark & 4.22 & 4.74 & 7.29 \\
    FocalClick  & SegFormer-B0-S1 & \cmark & 3.72 & 4.98 & 7.60 \\
    \rowcolor[gray]{0.9}
    SimpleClick & ViT-xTiny       & \xmark & 3.72 & 4.71 & 7.09 \\
    \bottomrule
  \end{tabular}
  \caption{\textbf{Comparison results on SBD for tiny models.} All models are trained on C+L with 230 epochs. Our SimpleClick-xTiny model outperforms FocalClick models without pretraining.}
  \label{tab:tiny_vits}
 }
\end{table}

\noindent\textbf{Computational Analysis}
Tab.~\ref{tab:computation_analysis} shows a comparison of computational requirements with respect to model parameters, FLOPs, GPU memory consumption, and speed; the speed is measured by seconds per click (SPC). Fig.~\ref{fig:teaser} shows the interactive segmentation performance of methods in terms of FLOPs. In Fig.~\ref{fig:teaser} and Tab.~\ref{tab:computation_analysis}, each method is denoted by its backbone.
For fair comparison, we evaluate all the methods on the same benchmark (\ie GrabCut) and using the same computer (GPU: NVIDIA RTX A6000, CPU: Intel Silver$\times$2).
We only calculate the FLOPs in a single forward pass. For methods like FocusCut which require multiple forward passes for each click, the FLOPs may be much higher than reported. By default, our method takes images of size 448$\times$448 as the fixed input. Even for our ViT-H model, the speed (132ms) and memory consumption (3.22G) is sufficient to meet the requirements of a practical annotation tool.

\begin{table}
\footnotesize
\centering
\begin{tabular}{l r r r r r}
    \toprule
    Backbone & Params/M & FLOPs/G & Mem/G & $\downarrow$SPC/ms \\
    \midrule
    HR-18s $_{400}$~\cite{sofiiuk2021reviving}      & 4.22   & 17.94  & 0.50  & 54  \\
    HR-18 $_{400}$~\cite{sofiiuk2021reviving}       & 10.03  & 30.99  & 0.52  & 56  \\
    HR-32 $_{400}$~\cite{sofiiuk2021reviving}       & 30.95  & 83.12  & 1.12  & 86  \\
    Swin-B $_{400}$~\cite{liu2022isegformer}        & 87.44  & 138.21 & 1.41  & 36  \\
    Swin-L $_{400}$~\cite{liu2022isegformer}        & 195.90 & 302.78 & 2.14  & 44  \\
    SegF-B0 $_{256}$~\cite{chen2022focalclick}      & 3.72   & 3.42   & 0.10  & 37  \\
    SegF-B3 $_{256}$~\cite{chen2022focalclick}      & 45.66  & 24.75  & 0.32  & 53  \\
    ResN-34 $_{384}$~\cite{chen2021conditional}     & 23.47  & 113.60 & 0.25  & 34  \\
    ResN-50 $_{384}$~\cite{lin2022focuscut}         & 40.36  & 78.82  & 0.85  & 331 \\ 
    ResN-101 $_{384}$~\cite{lin2022focuscut}        & 59.35  & 100.76 & 0.89  & 355 \\
    \midrule
    Ours-ViT-xT $_{224}$ &  3.72 &  2.63   & 0.17    & 17 \\
    Ours-ViT-xT $_{448}$ &  3.72 &  10.52  & 0.23    & 29 \\    
    Ours-ViT-B $_{224}$ & 96.46  & 42.44   & 0.51    & 34  \\
    Ours-ViT-B $_{448}$ & 96.46  & 169.78  & 0.87    & 54  \\
    Ours-ViT-L $_{448}$ & 322.18 & 532.87  & 1.72    & 86  \\
    Ours-ViT-H $_{448}$ & 659.39 & 1401.93 & 3.22    & 132 \\
    \bottomrule
\end{tabular}
\caption{\textbf{Computation comparison} for model parameters, FLOPs, GPU memory consumption (measured by the maximum GPU memory managed by PyTorch's caching allocator), and speed (measured by seconds per click). Each method is denoted by its backbone. The small number in front of the model denotes the input size ($448\times448$ for our models by default). Even for our ViT-H model, the speed (132ms) and memory consumption (3.22G) are sufficient to meet the requirements of a practical annotation tool.}
\label{tab:computation_analysis}
\end{table}




\begin{table}
\footnotesize
\centering
\begin{tabular}{l c c c c c}
    \toprule
    \multirow{2}{*}{FP design} & \multirow{2}{*}{frozen ViT} & \multicolumn{2}{c}{ViT-B} & \multicolumn{2}{c}{ViT-L} \\
    & & SBD & Pascal & SBD & Pascal \\ 
    \midrule
    (a) simple FP   & \cmark & 11.48 & 6.93 & 9.75 & 5.59 \\
    \rowcolor[gray]{0.9}    
    (a) simple FP   & \xmark & 5.24 & 2.53 & 4.46 & 2.15 \\
    (b) single-scale  & \xmark & 6.56 & 2.80 & 5.53 & 2.48 \\
    (c) parallel & \xmark & 7.21 & 3.09 & 6.26 & 2.79 \\
    (d) partial  & \xmark & 8.29 & 4.34 & 7.51 & 4.25 \\
    \bottomrule
\end{tabular}
\caption{\textbf{Ablation study} on backbone finetuning and feature pyramid (FP) design (Fig.~\ref{fig:ablation_feature_pyramid}). The metric is NoC@90. We have three findings in this ablation: 1) freezing the ViT backbone during finetuning significantly deteriorates the performance; 2) the multi-scale property matters for the simple feature pyramid; 3) the last feature map from the backbone is sufficient to build an effective feature pyramid.}
\label{tab:ablation_study}
\end{table}

\begin{figure*}[t]
    \centering
  \includegraphics[width=0.90\textwidth, height=2.3cm]{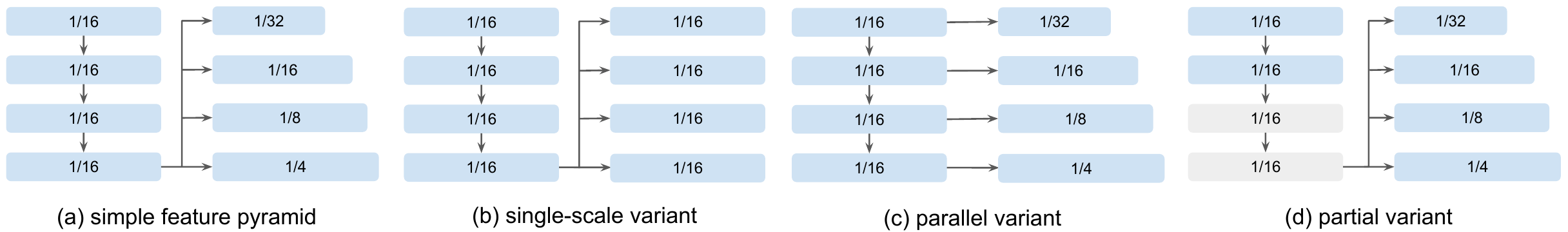}
  \caption{\textbf{Simple feature pyramid and its variants}: b) the single-scale variant ablates the multi-scale property; c) the parallel variant evenly extracts features from the backbone; d) the partial variant freezes the second half parameters of the backbone. Tab.~\ref{tab:ablation_study} shows the comparison results.}
  \label{fig:ablation_feature_pyramid}
\end{figure*}

\begin{figure*}[t]
    \centering
    \includegraphics[width=\textwidth, height=7.0cm]{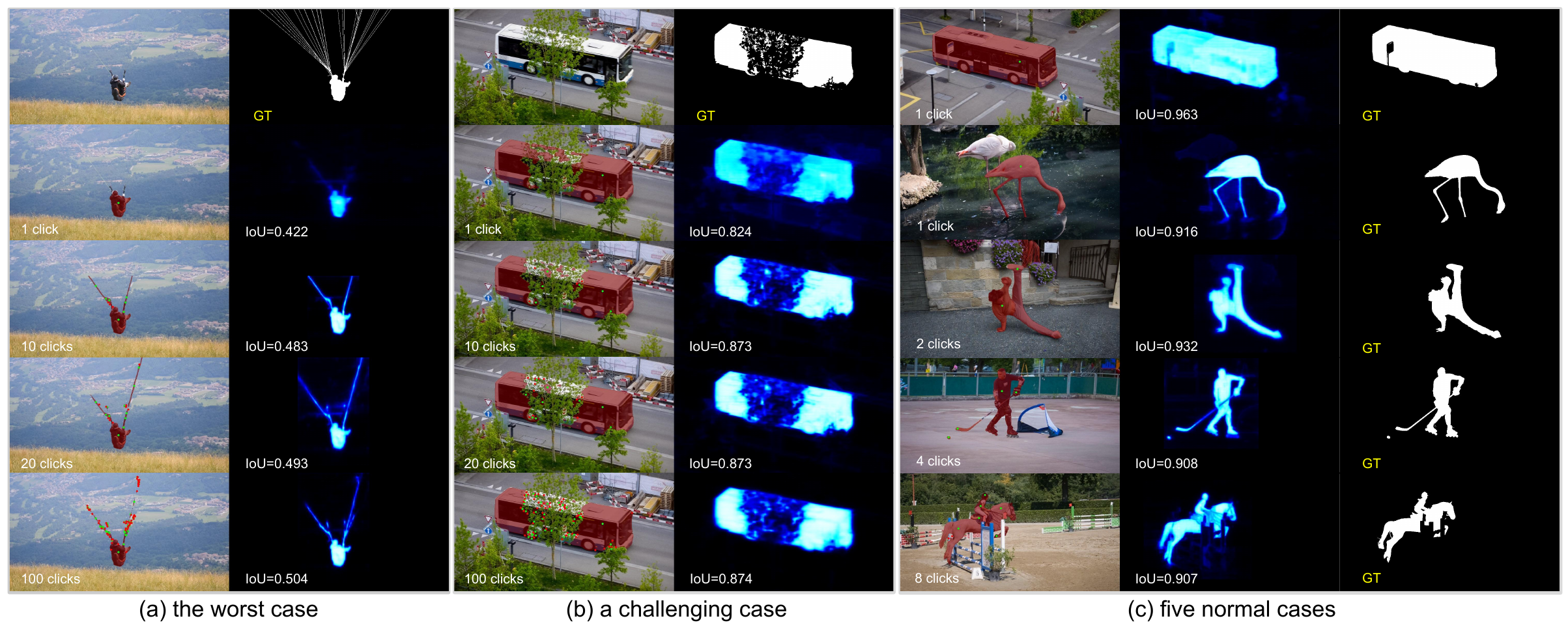}
   \caption{\textbf{Segmentation results} on DAVIS~\cite{perazzi2016benchmark}: (a) the worst case; (b) a challenging case; (c) five normal cases. The backbone model is ViT-L trained on COCO~\cite{lin2014microsoft}+LVIS~\cite{gupta2019lvis}. The segmentation probability maps are shown in blue; the segmentation maps are overlaid in red on the original images. The clicks are shown as green (positive click) or red (negative click) dots on the image. GT denotes ground truth.}
   \label{fig:qualitative-results}
\end{figure*}

\subsection{Ablation Study}
\label{sec:ablation_study}

In this section, we ablate the backbone finetuning and feature pyramid design.
Tab.~\ref{tab:ablation_study} shows the ablation results.
By default, we finetune the backbone along with other modules.
As an ablation, we freeze the backbone during finetuning, leading to significantly worse performance.
This ablation is explainable considering the ViT backbone takes most of the model parameters (Tab.~\ref{tab:model_size_comparison}).
For the second ablation, we compare the default simple feature pyramid design with three variants depicted in Fig.~\ref{fig:ablation_feature_pyramid} (\ie (b), (c), and (d)).
First, we observe that the multi-scale representation matters for the feature pyramid.
By ablating the multi-scale property in the simple feature pyramid, the performance drops considerably.
We also notice that the \emph{last} feature map from the backbone is strong enough to build the feature pyramid.
The parallel feature pyramid generated by multi-stage feature maps from the backbone does not surpass the simple feature pyramid that \emph{only} uses the last feature map of the backbone.

\section{Limitations and Remarks}
Our best-performing model (ViT-H) is much larger than existing models, leading to concerns about an unfair comparison. We justify the effectiveness of {\SimpleClick} by developing a tiny model and comparing it fairly with other methods. Other than this, our models may fail in some challenging scenarios such as objects with very thin and elongated shapes or cluttered occlusions((a) and (b) in Fig.~\ref{fig:qualitative-results}). We leave the improvements for future work.

We are entering an era of large-scale pretraining on multimodal foundation models, which is dramatically transforming the landscape of vision and language tasks. In this context, we hope SimpleClick will serve as a strong baseline for a new wave of high-performing interactive segmentation methods based on ViTs and large-scale pertaining.

\section{Conclusions}
\label{sec:conclusion}
We proposed {\SimpleClick}, the first plain-backbone method for interactive image segmentation. Our method leveraged a general-purpose ViT backbone that can benefit from readily available pretrained ViT models. With the MAE-pretrained weights, {\SimpleClick} achieved state-of-the-art performance on natural images and demonstrated strong generalizability on medical images. We also developed a tiny {\SimpleClick} model and provided a detailed computational analysis, highlighting the suitability of {\SimpleClick} as a practical annotation tool.
{\small
\bibliographystyle{ieee_fullname}
\bibliography{main}

\begin{thebibliography}{10}\itemsep=-1pt

\bibitem{acuna2018efficient}
David Acuna, Huan Ling, Amlan Kar, and Sanja Fidler.
\newblock Efficient interactive annotation of segmentation datasets with
  polygon-rnn++.
\newblock In {\em CVPR}, pages 859--868, 2018.

\bibitem{ambellan2019automated}
Felix Ambellan, Alexander Tack, Moritz Ehlke, and Stefan Zachow.
\newblock Automated segmentation of knee bone and cartilage combining
  statistical shape knowledge and convolutional neural networks: Data from the
  osteoarthritis initiative.
\newblock {\em Medical image analysis}, 52:109--118, 2019.

\bibitem{bai2007geodesic}
Xue Bai and Guillermo Sapiro.
\newblock A geodesic framework for fast interactive image and video
  segmentation and matting.
\newblock In {\em 2007 IEEE 11th International Conference on Computer Vision},
  pages 1--8. IEEE, 2007.

\bibitem{baid2021rsna}
Ujjwal Baid, Satyam Ghodasara, Suyash Mohan, Michel Bilello, Evan Calabrese,
  Errol Colak, Keyvan Farahani, Jayashree Kalpathy-Cramer, Felipe~C Kitamura,
  Sarthak Pati, et~al.
\newblock The rsna-asnr-miccai brats 2021 benchmark on brain tumor segmentation
  and radiogenomic classification.
\newblock {\em arXiv preprint arXiv:2107.02314}, 2021.

\bibitem{bertasius2020classifying}
Gedas Bertasius and Lorenzo Torresani.
\newblock Classifying, segmenting, and tracking object instances in video with
  mask propagation.
\newblock In {\em CVPR}, pages 9739--9748, 2020.

\bibitem{boykov2001interactive}
Yuri~Y Boykov and M-P Jolly.
\newblock Interactive graph cuts for optimal boundary \& region segmentation of
  objects in {ND} images.
\newblock In {\em Proceedings eighth IEEE international conference on computer
  vision. ICCV 2001}, volume~1, pages 105--112. IEEE, 2001.

\bibitem{caesar2020nuscenes}
Holger Caesar, Varun Bankiti, Alex~H Lang, Sourabh Vora, Venice~Erin Liong,
  Qiang Xu, Anush Krishnan, Yu Pan, Giancarlo Baldan, and Oscar Beijbom.
\newblock nuscenes: A multimodal dataset for autonomous driving.
\newblock In {\em CVPR}, pages 11621--11631, 2020.

\bibitem{chen2021simple}
Wuyang Chen, Xianzhi Du, Fan Yang, Lucas Beyer, Xiaohua Zhai, Tsung-Yi Lin,
  Huizhong Chen, Jing Li, Xiaodan Song, Zhangyang Wang, et~al.
\newblock A simple single-scale vision transformer for object localization and
  instance segmentation.
\newblock {\em arXiv preprint arXiv:2112.09747}, 2021.

\bibitem{chen2021conditional}
Xi Chen, Zhiyan Zhao, Feiwu Yu, Yilei Zhang, and Manni Duan.
\newblock Conditional diffusion for interactive segmentation.
\newblock In {\em ICCV}, pages 7345--7354, 2021.

\bibitem{chen2022focalclick}
Xi Chen, Zhiyan Zhao, Yilei Zhang, Manni Duan, Donglian Qi, and Hengshuang
  Zhao.
\newblock {FocalClick}: Towards practical interactive image segmentation.
\newblock In {\em CVPR}, pages 1300--1309, 2022.

\bibitem{deng2009imagenet}
Jia Deng, Wei Dong, Richard Socher, Li-Jia Li, Kai Li, and Li Fei-Fei.
\newblock Imagenet: A large-scale hierarchical image database.
\newblock In {\em 2009 IEEE conference on computer vision and pattern
  recognition}, pages 248--255. Ieee, 2009.

\bibitem{ding2021local}
Zhipeng Ding, Xu Han, Peirong Liu, and Marc Niethammer.
\newblock Local temperature scaling for probability calibration.
\newblock In {\em ICCV}, pages 6889--6899, 2021.

\bibitem{dosovitskiy2020image}
Alexey Dosovitskiy, Lucas Beyer, Alexander Kolesnikov, Dirk Weissenborn,
  Xiaohua Zhai, Thomas Unterthiner, Mostafa Dehghani, Matthias Minderer, Georg
  Heigold, Sylvain Gelly, et~al.
\newblock An image is worth 16x16 words: Transformers for image recognition at
  scale.
\newblock {\em arXiv preprint arXiv:2010.11929}, 2020.

\bibitem{everingham2010pascal}
Mark Everingham, Luc Van~Gool, Christopher~KI Williams, John Winn, and Andrew
  Zisserman.
\newblock The pascal visual object classes (voc) challenge.
\newblock {\em International journal of computer vision}, 88(2):303--338, 2010.

\bibitem{gerhard2013segmented}
Stephan Gerhard, Jan Funke, Julien Martel, Albert Cardona, and Richard Fetter.
\newblock Segmented anisotropic sstem dataset of neural tissue.
\newblock {\em figshare}, pages 0--0, 2013.

\bibitem{grady2006random}
Leo Grady.
\newblock Random walks for image segmentation.
\newblock {\em IEEE transactions on pattern analysis and machine intelligence},
  28(11):1768--1783, 2006.

\bibitem{gu2022multi}
Jiaqi Gu, Hyoukjun Kwon, Dilin Wang, Wei Ye, Meng Li, Yu-Hsin Chen, Liangzhen
  Lai, Vikas Chandra, and David~Z Pan.
\newblock Multi-scale high-resolution vision transformer for semantic
  segmentation.
\newblock In {\em CVPR}, pages 12094--12103, 2022.

\bibitem{gulshan2010geodesic}
Varun Gulshan, Carsten Rother, Antonio Criminisi, Andrew Blake, and Andrew
  Zisserman.
\newblock Geodesic star convexity for interactive image segmentation.
\newblock In {\em 2010 IEEE Computer Society Conference on Computer Vision and
  Pattern Recognition}, pages 3129--3136. IEEE, 2010.

\bibitem{gupta2019lvis}
Agrim Gupta, Piotr Dollar, and Ross Girshick.
\newblock Lvis: A dataset for large vocabulary instance segmentation.
\newblock In {\em CVPR}, pages 5356--5364, 2019.

\bibitem{hariharan2011semantic}
Bharath Hariharan, Pablo Arbel{\'a}ez, Lubomir Bourdev, Subhransu Maji, and
  Jitendra Malik.
\newblock Semantic contours from inverse detectors.
\newblock In {\em 2011 international conference on computer vision}, pages
  991--998. IEEE, 2011.

\bibitem{he2021masked}
Kaiming He, Xinlei Chen, Saining Xie, Yanghao Li, Piotr Doll{\'a}r, and Ross
  Girshick.
\newblock Masked autoencoders are scalable vision learners.
\newblock {\em arXiv preprint arXiv:2111.06377}, 2021.

\bibitem{he2016deep}
Kaiming He, Xiangyu Zhang, Shaoqing Ren, and Jian Sun.
\newblock Deep residual learning for image recognition.
\newblock In {\em CVPR}, pages 770--778, 2016.

\bibitem{jang2019interactive}
Won-Dong Jang and Chang-Su Kim.
\newblock Interactive image segmentation via backpropagating refinement scheme.
\newblock In {\em CVPR}, pages 5297--5306, 2019.

\bibitem{khan2022transformers}
Salman Khan, Muzammal Naseer, Munawar Hayat, Syed~Waqas Zamir, Fahad~Shahbaz
  Khan, and Mubarak Shah.
\newblock Transformers in vision: A survey.
\newblock {\em ACM computing surveys (CSUR)}, 54(10s):1--41, 2022.

\bibitem{li2022exploring}
Yanghao Li, Hanzi Mao, Ross Girshick, and Kaiming He.
\newblock Exploring plain vision transformer backbones for object detection.
\newblock {\em arXiv preprint arXiv:2203.16527}, 2022.

\bibitem{li2018interactive}
Zhuwen Li, Qifeng Chen, and Vladlen Koltun.
\newblock Interactive image segmentation with latent diversity.
\newblock In {\em CVPR}, pages 577--585, 2018.

\bibitem{lin2017feature}
Tsung-Yi Lin, Piotr Doll{\'a}r, Ross Girshick, Kaiming He, Bharath Hariharan,
  and Serge Belongie.
\newblock Feature pyramid networks for object detection.
\newblock In {\em CVPR}, pages 2117--2125, 2017.

\bibitem{lin2014microsoft}
Tsung-Yi Lin, Michael Maire, Serge Belongie, James Hays, Pietro Perona, Deva
  Ramanan, Piotr Doll{\'a}r, and C~Lawrence Zitnick.
\newblock Microsoft coco: Common objects in context.
\newblock In {\em ECCV}, pages 740--755. Springer, 2014.

\bibitem{lin2022focuscut}
Zheng Lin, Zheng-Peng Duan, Zhao Zhang, Chun-Le Guo, and Ming-Ming Cheng.
\newblock {FocusCut}: Diving into a focus view in interactive segmentation.
\newblock In {\em CVPR}, pages 2637--2646, 2022.

\bibitem{lin2020interactive}
Zheng Lin, Zhao Zhang, Lin-Zhuo Chen, Ming-Ming Cheng, and Shao-Ping Lu.
\newblock Interactive image segmentation with first click attention.
\newblock In {\em CVPR}, pages 13339--13348, 2020.

\bibitem{litjens2017survey}
Geert Litjens, Thijs Kooi, Babak~Ehteshami Bejnordi, Arnaud Arindra~Adiyoso
  Setio, Francesco Ciompi, Mohsen Ghafoorian, Jeroen~Awm Van Der~Laak, Bram
  Van~Ginneken, and Clara~I S{\'a}nchez.
\newblock A survey on deep learning in medical image analysis.
\newblock {\em Medical image analysis}, 42:60--88, 2017.

\bibitem{liu2022isegformer}
Qin Liu, Zhenlin Xu, Yining Jiao, and Marc Niethammer.
\newblock {iSegFormer}: Interactive segmentation via transformers with
  application to 3d knee mr images.
\newblock In {\em International Conference on Medical Image Computing and
  Computer-Assisted Intervention}, pages 464--474. Springer, 2022.

\bibitem{liu2022pseudoclick}
Qin Liu, Meng Zheng, Benjamin Planche, Srikrishna Karanam, Terrence Chen, Marc
  Niethammer, and Ziyan Wu.
\newblock {PseudoClick}: Interactive image segmentation with click imitation.
\newblock {\em arXiv preprint arXiv:2207.05282}, 2022.

\bibitem{liu2021swin}
Ze Liu, Yutong Lin, Yue Cao, Han Hu, Yixuan Wei, Zheng Zhang, Stephen Lin, and
  Baining Guo.
\newblock Swin transformer: Hierarchical vision transformer using shifted
  windows.
\newblock In {\em ICCV}, pages 10012--10022, 2021.

\bibitem{maninis2018deep}
Kevis-Kokitsi Maninis, Sergi Caelles, Jordi Pont-Tuset, and Luc Van~Gool.
\newblock Deep extreme cut: From extreme points to object segmentation.
\newblock In {\em CVPR}, pages 616--625, 2018.

\bibitem{martin2001database}
David Martin, Charless Fowlkes, Doron Tal, and Jitendra Malik.
\newblock A database of human segmented natural images and its application to
  evaluating segmentation algorithms and measuring ecological statistics.
\newblock In {\em Proceedings Eighth IEEE International Conference on Computer
  Vision. ICCV 2001}, volume~2, pages 416--423. IEEE, 2001.

\bibitem{paszke2019pytorch}
Adam Paszke, Sam Gross, Francisco Massa, Adam Lerer, James Bradbury, Gregory
  Chanan, Trevor Killeen, Zeming Lin, Natalia Gimelshein, Luca Antiga, et~al.
\newblock Pytorch: An imperative style, high-performance deep learning library.
\newblock {\em Advances in neural information processing systems}, 32, 2019.

\bibitem{perazzi2016benchmark}
Federico Perazzi, Jordi Pont-Tuset, Brian McWilliams, Luc Van~Gool, Markus
  Gross, and Alexander Sorkine-Hornung.
\newblock A benchmark dataset and evaluation methodology for video object
  segmentation.
\newblock In {\em CVPR}, pages 724--732, 2016.

\bibitem{rother2004grabcut}
Carsten Rother, Vladimir Kolmogorov, and Andrew Blake.
\newblock " grabcut" interactive foreground extraction using iterated graph
  cuts.
\newblock {\em ACM transactions on graphics (TOG)}, 23(3):309--314, 2004.

\bibitem{shen2017deep}
Dinggang Shen, Guorong Wu, and Heung-Il Suk.
\newblock Deep learning in medical image analysis.
\newblock {\em Annual review of biomedical engineering}, 19:221, 2017.

\bibitem{sofiiuk2020f}
Konstantin Sofiiuk, Ilia Petrov, Olga Barinova, and Anton Konushin.
\newblock f-brs: Rethinking backpropagating refinement for interactive
  segmentation.
\newblock In {\em CVPR}, pages 8623--8632, 2020.

\bibitem{sofiiuk2021reviving}
Konstantin Sofiiuk, Ilia~A Petrov, and Anton Konushin.
\newblock Reviving iterative training with mask guidance for interactive
  segmentation.
\newblock {\em arXiv preprint arXiv:2102.06583}, 2021.

\bibitem{strudel2021segmenter}
Robin Strudel, Ricardo Garcia, Ivan Laptev, and Cordelia Schmid.
\newblock Segmenter: Transformer for semantic segmentation.
\newblock In {\em ICCV}, pages 7262--7272, 2021.

\bibitem{wu2014milcut}
Jiajun Wu, Yibiao Zhao, Jun-Yan Zhu, Siwei Luo, and Zhuowen Tu.
\newblock Milcut: A sweeping line multiple instance learning paradigm for
  interactive image segmentation.
\newblock In {\em CVPR}, pages 256--263, 2014.

\bibitem{xie2021segformer}
Enze Xie, Wenhai Wang, Zhiding Yu, Anima Anandkumar, Jose~M Alvarez, and Ping
  Luo.
\newblock {SegFormer}: Simple and efficient design for semantic segmentation
  with transformers.
\newblock {\em Advances in Neural Information Processing Systems},
  34:12077--12090, 2021.

\bibitem{xu2017deep}
Ning Xu, Brian Price, Scott Cohen, Jimei Yang, and Thomas Huang.
\newblock Deep grabcut for object selection.
\newblock {\em arXiv preprint arXiv:1707.00243}, 2017.

\bibitem{xu2016deep}
Ning Xu, Brian Price, Scott Cohen, Jimei Yang, and Thomas~S Huang.
\newblock Deep interactive object selection.
\newblock In {\em CVPR}, pages 373--381, 2016.

\bibitem{xu2018youtube}
Ning Xu, Linjie Yang, Yuchen Fan, Dingcheng Yue, Yuchen Liang, Jianchao Yang,
  and Thomas Huang.
\newblock Youtube-vos: A large-scale video object segmentation benchmark.
\newblock {\em arXiv preprint arXiv:1809.03327}, 2018.

\bibitem{yuan2021hrformer}
Yuhui Yuan, Rao Fu, Lang Huang, Weihong Lin, Chao Zhang, Xilin Chen, and
  Jingdong Wang.
\newblock Hrformer: High-resolution vision transformer for dense predict.
\newblock {\em Advances in Neural Information Processing Systems},
  34:7281--7293, 2021.

\bibitem{zhang2020interactive}
Shiyin Zhang, Jun~Hao Liew, Yunchao Wei, Shikui Wei, and Yao Zhao.
\newblock Interactive object segmentation with inside-outside guidance.
\newblock In {\em CVPR}, pages 12234--12244, 2020.

\bibitem{zheng2021rethinking}
Sixiao Zheng, Jiachen Lu, Hengshuang Zhao, Xiatian Zhu, Zekun Luo, Yabiao Wang,
  Yanwei Fu, Jianfeng Feng, Tao Xiang, Philip~HS Torr, et~al.
\newblock Rethinking semantic segmentation from a sequence-to-sequence
  perspective with transformers.
\newblock In {\em CVPR}, pages 6881--6890, 2021.

\end{thebibliography}
}

\clearpage
\begin{appendices}

\section{Datasets}
\label{supp:datasets}

\textcolor{blue}{This section supplements the ``Datasets'' (Sec.~\ref{sec:experiments}) in the main paper.} Our models are trained either using SBD~\cite{hariharan2011semantic} or the combined COCO~\cite{lin2014microsoft}+LVIS~\cite{gupta2019lvis} datasets. Before RITM~\cite{sofiiuk2021reviving}, most of the deep learning-based interactive segmentation models were trained either using the SBD~\cite{hariharan2011semantic} or Pascal VOC~\cite{everingham2010pascal} datasets. These two datasets only cover 20 categories of general objects such as persons, transportation vehicles, animals, and indoor objects. The authors of RITM constructed the combined COCO+LVIS dataset, which contains 118k training images of 80 diverse object classes, for interactive segmentation. This large and diverse training dataset contributes to the state-of-the-art performance of RITM models. Inspired by RITM and its follow-up works~\cite{chen2022focalclick,liu2022pseudoclick}, we use SBD and COCO+LVIS as our training datasets.

\section{Implementation Details}
\subsection{Architectures}
\textcolor{blue}{This section supplements Sec.~\ref{sec:method_network_architecture} ``Network Architecture" in the main paper.}
Tab.~\ref{tab:architecture_parameters} shows the main architecture parameters of our models. By default, our models use an input size of $448\times448$ during training and evaluation.
Our ViT-B and ViT-L models use a patch size of $16\times16$, while the ViT-H model uses a smaller patch size of $14\times14$. This leads to a higher resolution representation in terms of the number of patches. Each patch is flattened and projected to an embed dimension of $C_0$ through the patch embedding layer. The tokens generated by the patch embedding layer are processed by $N$ self-attention blocks, which $N$ is a hyper-parameter inherited from plain ViT models~\cite{he2021masked}. Inspired by ViTDet~\cite{li2022exploring}, we build a simple feature pyramid with the four resolutions \{$\frac{1}{32}, \frac{1}{16}, \frac{1}{8}, \frac{1}{4}$\}. The $\frac{1}{16}$ resolution uses the last feature map of the ViT backbone. The $\frac{1}{32}$ resolution is built by a $2\times2$ convolutional layer with a stride of 2. The $\frac{1}{8}$ (or $\frac{1}{4}$) resolution is built by one (or two) $2\times2$ transposed convolution layer(s) with a stride of 2. We use a $1\times1$ convolution layer with layer normalization to convert the channels of each feature map to predefined dimensions.  Specifically, feature maps of resolutions \{$\frac{1}{32}, \frac{1}{16}, \frac{1}{8}, \frac{1}{4}$\} are converted to channel dimensions of \{$8C_1, 4C_1, 2C_1, C_1$\}, respectively. Each feature map is then converted to the same dimension of $C_2$ through an MLP layer in the segmentation head, followed by upsampling to the $\frac{1}{4}$ resolution. At this point, the four feature maps have the same resolution and the same number of channels. They are concatenated as a single feature map with $4C_2$ channels. Another MLP layer in the segmentation head converts this multi-channel feature map to a one-channel feature map, followed by a sigmoid function to obtain the final binary segmentation. We use $C_1$ and $C_2$ as hyper-parameters without tuning.

\begin{table}[h]
\footnotesize
\renewcommand\arraystretch{0.9}
  \centering
  \begin{tabular}{l c c c c}
    \toprule
    Model & $H,W$ & Patch Size & $N$ & $C_0$, $C_1$, $C_2$ \\
    \midrule
    Ours-ViT-B & 448, 448 & $16\times16$ & 12 & 768, 128, 256 \\
    Ours-ViT-L & 448, 448 & $16\times16$ & 24 & 1024, 192, 256 \\
    Ours-ViT-H & 448, 448 & $14\times14$ & 32 & 1280, 240, 256 \\
    \bottomrule
  \end{tabular}
  \caption{\textbf{Architecture parameters} of {\SimpleClick} models. $N$ denotes the number of self-attention blocks. $C_0$, $C_1$, and $C_2$ denote the feature map dimensions at different levels.} 
  \label{tab:architecture_parameters}
\end{table}

\subsection{Clicks Encoding}
\textcolor{blue}{This section also supplements Sec.~\ref{sec:method_network_architecture} in the main paper.}
We encode clicks, which are represented by the coordinates in an image, as disks with a small radius of 5 pixels. Positive and negative clicks are encoded separately. In our implementation, we also attach the previous segmentation as an additional channel, resulting in a three-channel disk map. Two patch embedding layers, which are of the same structure, process the three-channel disk map and the RGB image separately. The tokens of the two inputs after the patch embedding layers are added element by element, without changing the input dimensions for the self-attention blocks. This design is more efficient than other designs such as concatenation and allows our ViT backbones to be initialized with pretrained ViT weights.

\subsection{Finetuning on Higher-Resolution Images}
\textcolor{blue}{This section supplements Sec.~\ref{sec:method_training_inference} ``Training and Inference Settings''in the main paper.}
Our models are pretrained on an image size of $224\times224$ but are finetuned on an image size of $448\times448$. We first interpolate the positional encoding to the high resolution. Then, we perform non-overlapping window attention~\cite{li2022exploring} with a few global blocks for cross-window attention. The high-resolution feature map is divided into regular non-overlapping windows. The non-global blocks perform self-attention within each window, while global blocks perform global self-attention. We set the number of global blocks to 2, 6, and 8 for the ViT-B, ViT-L, and ViT-H models, respectively.

\section{Additional Comparison Results}
\label{supp:additional-comparison-results}

\textcolor{blue}{This section supplements Sec.~\ref{sec:comparison-with-previous-results} ``Comparison with Previous Results'' in the main paper.} Fig.~\ref{fig:convergence_analysis} shows convergence results for our models on four datasets: GrabCut~\cite{rother2004grabcut}, Berkeley~\cite{martin2001database}, DAVIS~\cite{perazzi2016benchmark}, and COCO~\cite{lin2014microsoft}. Overall, our models perform better than other models on these datasets. However, the results in Fig.~\ref{fig:convergence_analysis} are not as compelling as the results on SBD~\cite{hariharan2011semantic} or Pascal VOC~\cite{everingham2010pascal} (shown in Fig. 3 of the main paper). This is likely due to the limited number of images in these datasets (\eg GrabCut only contains 50 instances, while SBD contains 6671 instances for evaluation). 

\begin{figure*}[t]
    \includegraphics[width=\textwidth, height=19.0cm]{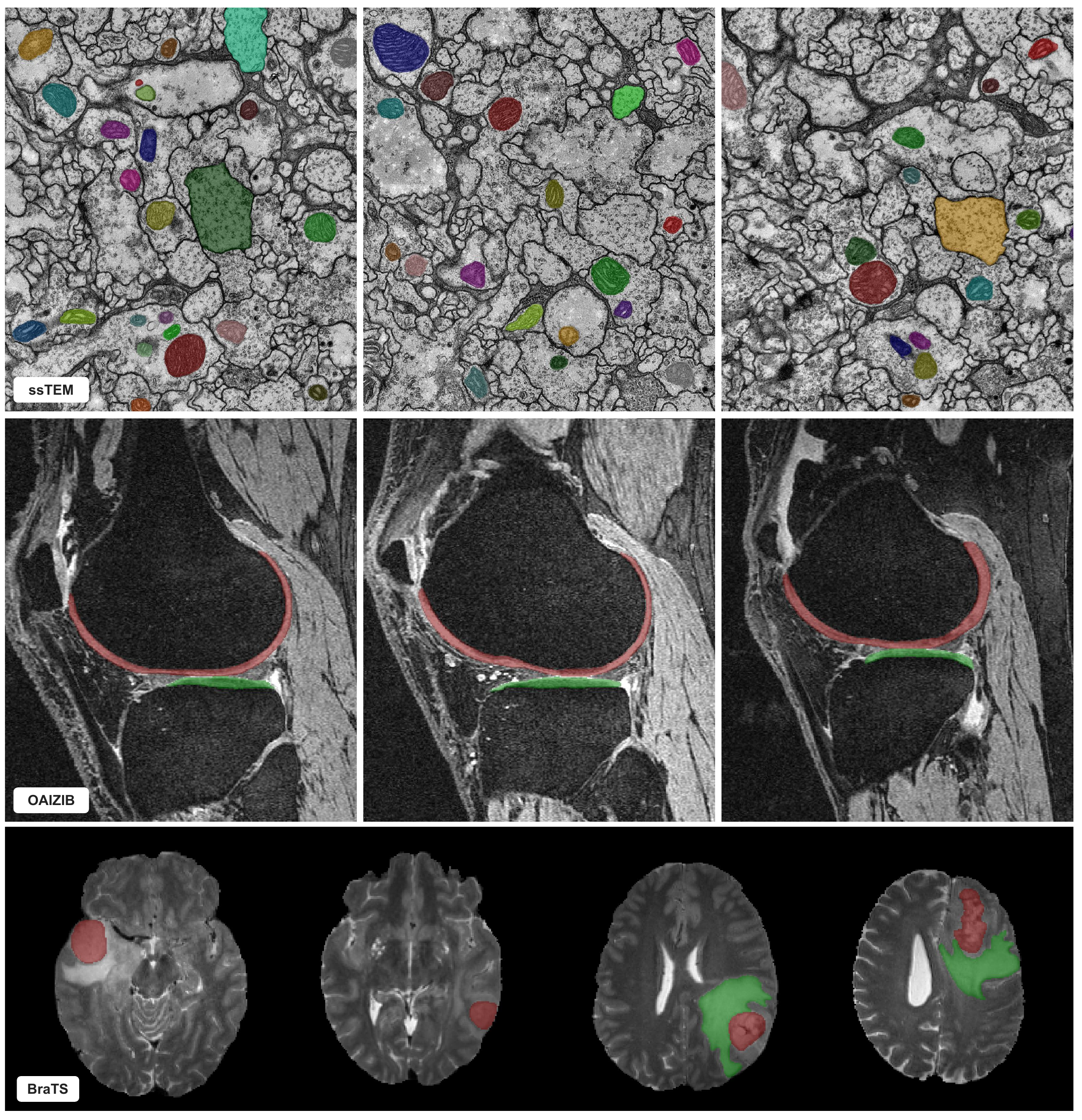}
   \caption{\textbf{Qualitative results} of human evaluation on three medical image datasets: ssTEM~\cite{gerhard2013segmented}, OAIZIB~\cite{ambellan2019automated}, and BraTS~\cite{baid2021rsna}. All the results are obtained by a human-in-the-loop providing the clicks. Though our models are evaluated on medical images without finetuning, they generalize well to all the unseen objects given a few clicks, as shown in the demo videos (\href{https://github.com/uncbiag/SimpleClick}{https://github.com/uncbiag/SimpleClick}).}
   \label{fig:qualitative_evaluation}
\end{figure*}

\begin{figure*}[t!]
        \includegraphics[width=9.0cm, height=6.0cm, trim=40 5 40 5, clip]{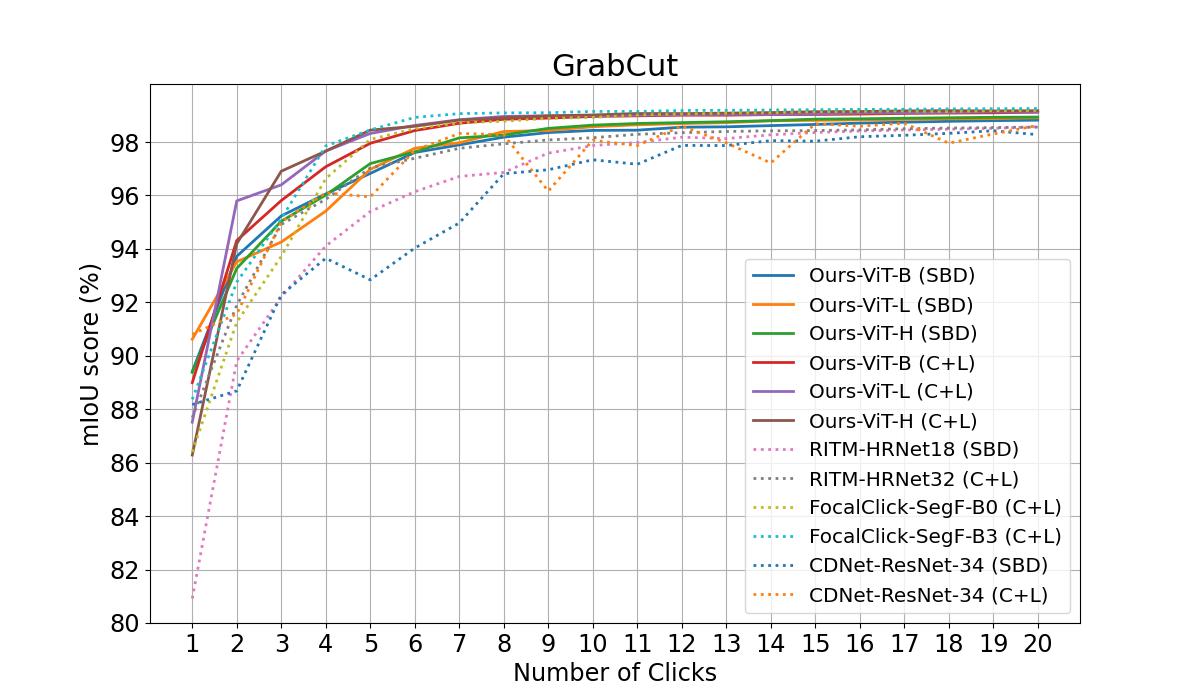}
        \includegraphics[width=9.0cm, height=6.0cm, trim=40 5 40 5, clip]{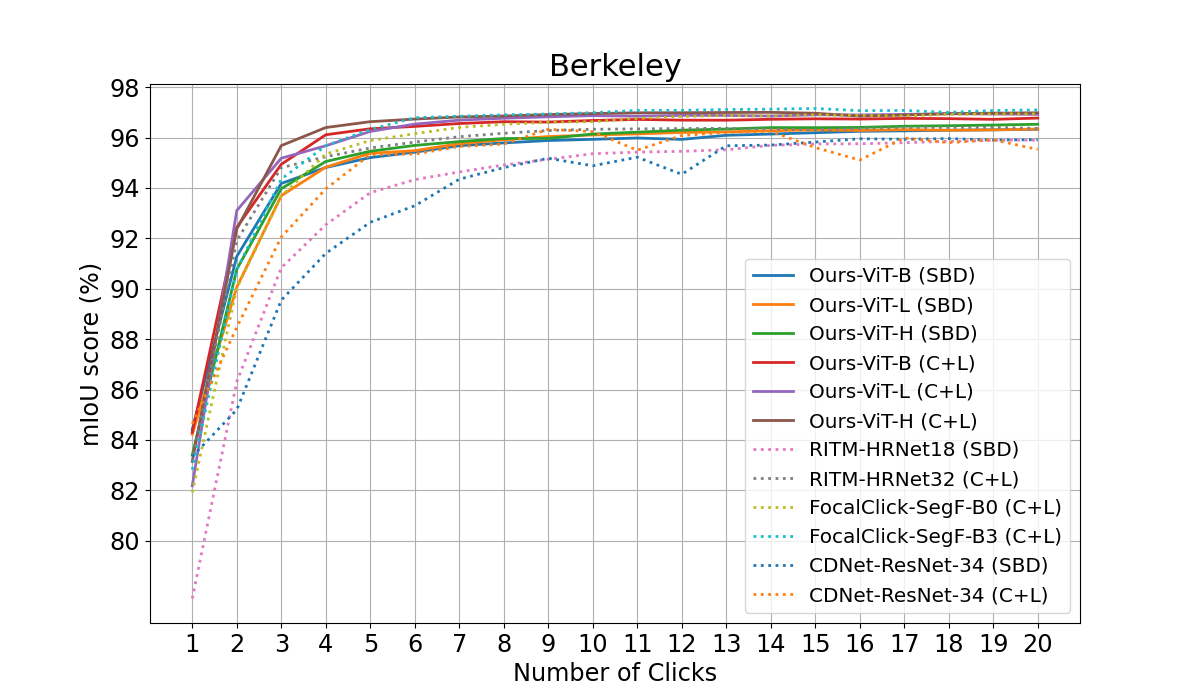}
        \includegraphics[width=9.0cm, height=6.0cm, trim=40 5 40 5, clip]{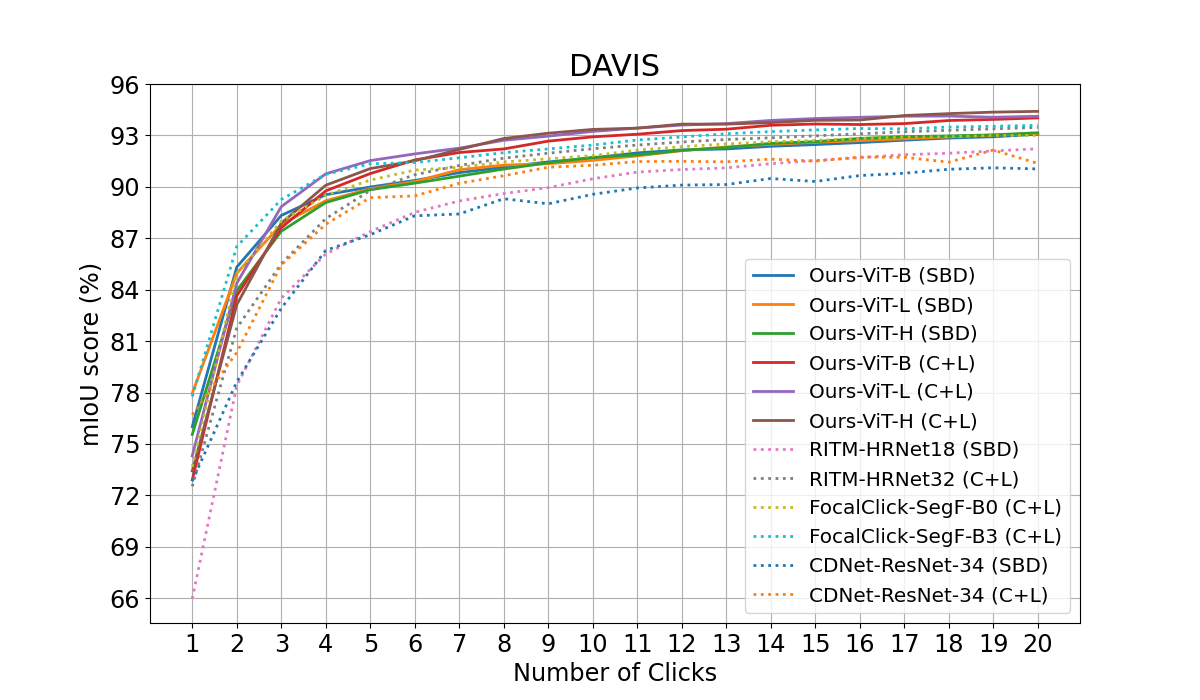}
        \includegraphics[width=9.0cm, height=6.0cm, trim=40 5 40 5, clip]{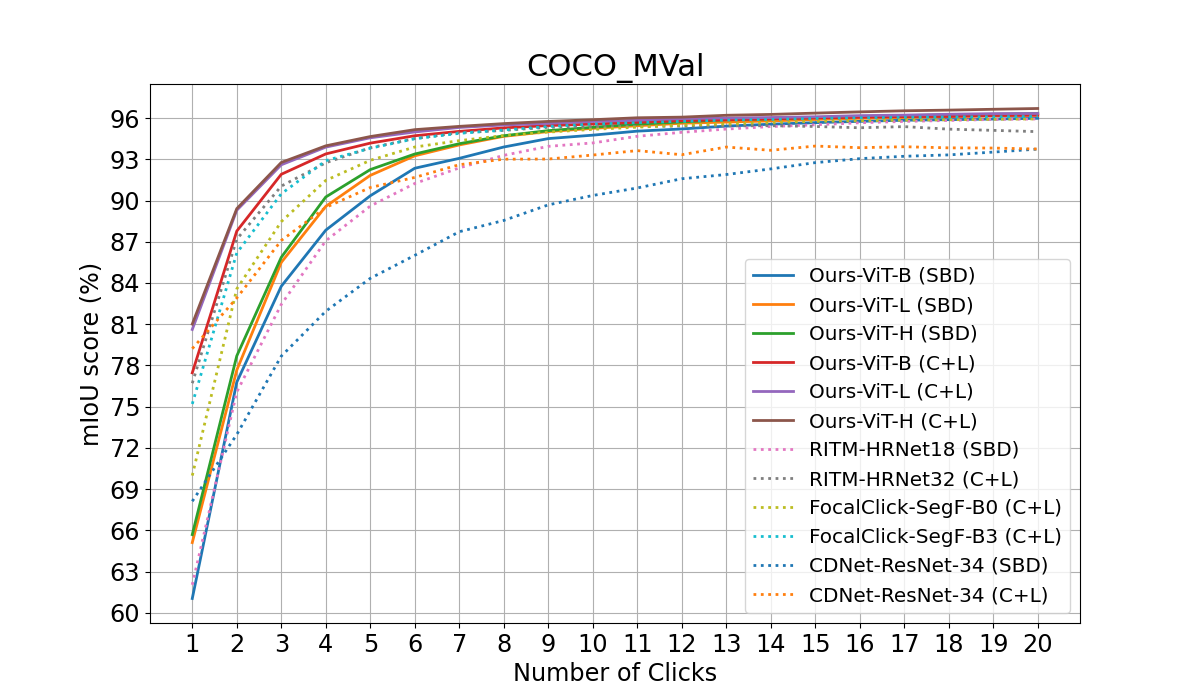}
    \caption{\textbf{Convergence analysis} for GrabCut, Berkeley, DAVIS, and COCO. All models are trained on either SBD~\cite{hariharan2011semantic} or COCO~\cite{lin2014microsoft}+LVIS~\cite{gupta2019lvis} (C+L). The metric is mean IoU given $k$ clicks (mIoU@$k$). In general, our models require fewer clicks for a given accuracy level.}
    \label{fig:convergence_analysis}
\end{figure*}

\section{Human Evaluation on Medical Images}
\textcolor{blue}{This section supplemens Sec.~\ref{sec:OOD-evaluation-on-medical-images} ``Out-of-Domain Evaluation on Medical Images" in the main paper.}
In the main paper, we report quantitative results on medical images using an automatic evaluation mode where clicks are automatically simulated. In this section, we perform human evaluations where a human-in-the-loop provides all the clicks. Fig.~\ref{fig:qualitative_evaluation} shows qualitative results on three medical image datasets: ssTEM~\cite{gerhard2013segmented}, OAIZIB~\cite{ambellan2019automated}, and BraTS~\cite{bai2007geodesic}. For simple objects such as cell nuclei in ssTEM, it may take as little as one click for a good segmentation. However, for more challenging objects such as knee cartilage in the OAIZIB dataset or brain tumors in the BraTS dataset, it may take more than ten clicks until a high-quality segmentation is obtained. Considering our models are not finetuned on the label-scarce medical imaging datasets, our observed performance is quite promising. The attached videos demonstrate the evaluation process.

\end{appendices}

\end{document}